%% file: root.tex
\documentclass[letterpaper, 10 pt, conference]{format/ieeeconf}  

\IEEEoverridecommandlockouts                              

\usepackage{amsmath} 
\usepackage{amssymb}  
\usepackage{xcolor}
\usepackage[ruled,vlined,linesnumbered]{algorithm2e}
\usepackage{graphicx}
\usepackage{lipsum}
\usepackage[noadjust]{cite}
\usepackage{subcaption}
\usepackage{float}
\usepackage{wrapfig}
\usepackage{pifont}

\newcommand{\rulesep}{\unskip\ \vrule\ }

\title{\bf Improving Kinodynamic Planners for Vehicular Navigation \\
with Learned Goal-Reaching Controllers}

\author{Aravind Sivaramakrishnan, Edgar Granados, Seth Karten, Troy McMahon, Kostas E. Bekris%
\thanks{The authors are with the Dept. of Computer Science at Rutgers, NJ, USA. Email: {\tt\small { \{aravind.siva, eg585, sak295, tm799, kb572\}}@rutgers.edu}. This work was supported by NSF HDR TRIPODS award 1934924. The results do not reflect the sponsor's positions.}%
}

\begin{document}

\maketitle
\thispagestyle{empty}
\pagestyle{empty}

\begin{abstract}

This paper aims to improve the path quality and computational efficiency of sampling-based kinodynamic planners for vehicular navigation. It proposes a learning framework for identifying promising controls during the expansion process of sampling-based planners. Given a dynamics model, a reinforcement learning process is trained offline to return a low-cost control that reaches a local goal state (i.e., a waypoint) in the absence of obstacles. By focusing on the system's dynamics and not knowing the environment, this process is data-efficient and takes place once for a robotic system. In this way, it can be reused in different environments. The planner generates online local goal states for the learned controller in an informed manner to bias towards the goal and consecutively in an exploratory, random manner. For the informed expansion, local goal states are generated either via (a) medial axis information in environments with obstacles, or (b) wavefront information for setups with traversability costs. The learning process and the resulting planning framework are evaluated for a first and second-order differential drive system, as well as a physically simulated Segway robot. The results show that the proposed integration of learning and planning can produce higher quality paths than sampling-based kinodynamic planning with random controls in fewer iterations and computation time. 
\end{abstract}

\input{sections/01_introduction}

\input{sections/02_problem}

\input{sections/03_framework}

\input{sections/04_method}

\input{sections/05_results}

\input{sections/06_conclusion}

\bibliographystyle{format/IEEEtran}
\bibliography{root.bib}

\end{document}

%% file: sections/01_introduction.tex
\section{Introduction}
\label{sec:introduction}

This work focuses on improving the efficiency of sampling-based motion planning for vehicular systems that exhibit significant dynamics using machine learning. Planning motions for such systems can be challenging when there is no access to a local planner and the basic primitive to explore the state space is forward propagation of controls. In this context, tree sampling-based kinodynamic planners have been developed \cite{lavalle2001randomized}, some of which achieve asymptotic optimality (AO) by propagating randomly sampled controls during each iteration \cite{li2016asymptotically, hauser2016asymptotically, LB-DIRT, kleinbort2020refined}. While random controls are desirable for analysis purposes, in practice they result in low quality (i.e., high cost) trajectories.  

Learning candidate controls for edge propagation from previous planning experience is a promising avenue for improving the practical efficiency of sampling-based kinodynamic planners. Prior work on the integration of machine learning for planning has primarily focused on geometric motion planning problems that don't deal with dynamics \cite{Qureshi2021MotionPN, jurgenson2019harnessing}, or requires access to a two-point boundary value problem (BVP) solver to construct a supervised learning dataset \cite{Li2018NeuralNA}. The latter may not be easily available for some vehicular models, especially for higher-fidelity physics-based models.  

Recent work has shown that deep reinforcement learning (DRL) can be used to learn a local controller (or a \textit{policy}) that steers the robot between two collision-free states. The learned policy can be then used as an edge connection strategy for a sampling-based motion planner, such as a Probabilistic Roadmap (PRM) \cite{Faust2018PRMRLLR} or a Rapidly-exploring Random Tree (RRT) \cite{chiang2019rl}. This approach, however, requires a large amount of training data and manual specification of reward parameters.  

\begin{figure}[t!]
    \centering
    \begin{subfigure}{\linewidth}
    \includegraphics[width=0.49\linewidth]{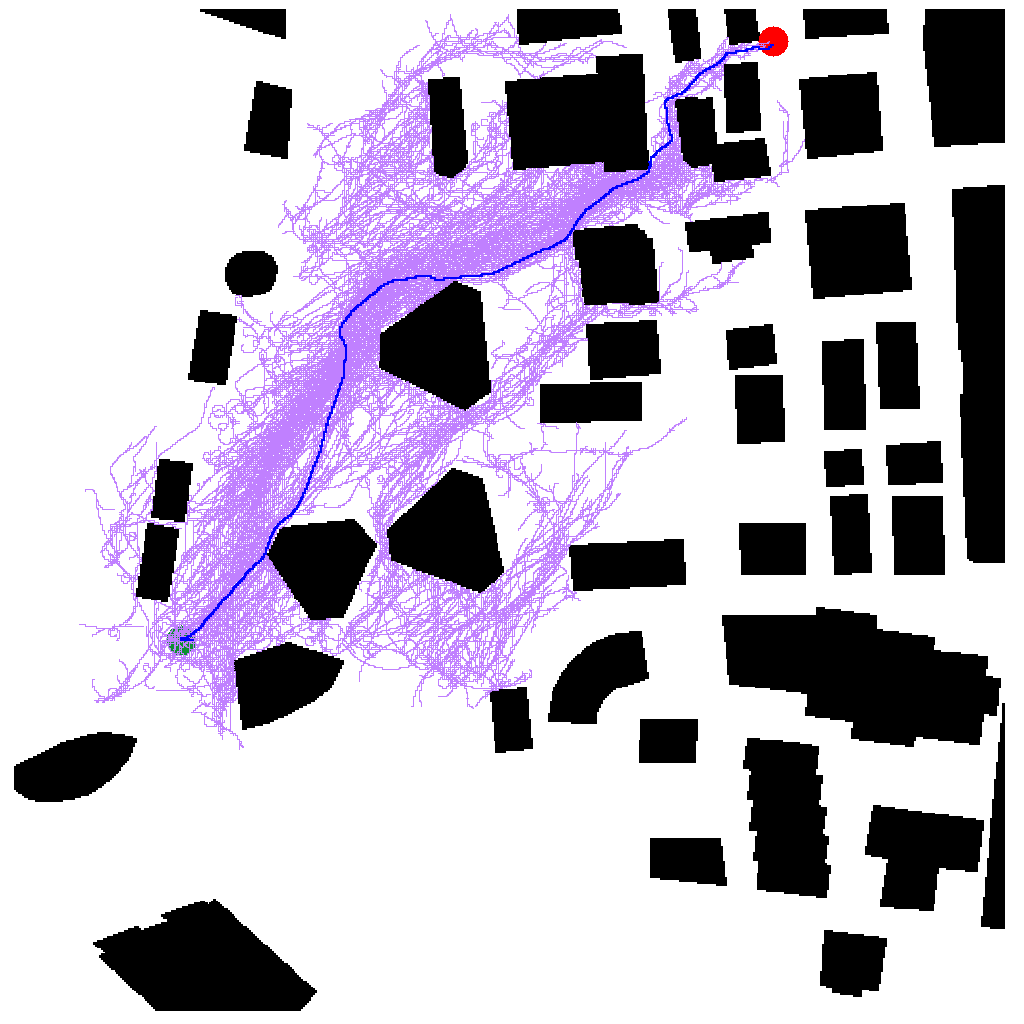}
    \includegraphics[width=0.49\linewidth]{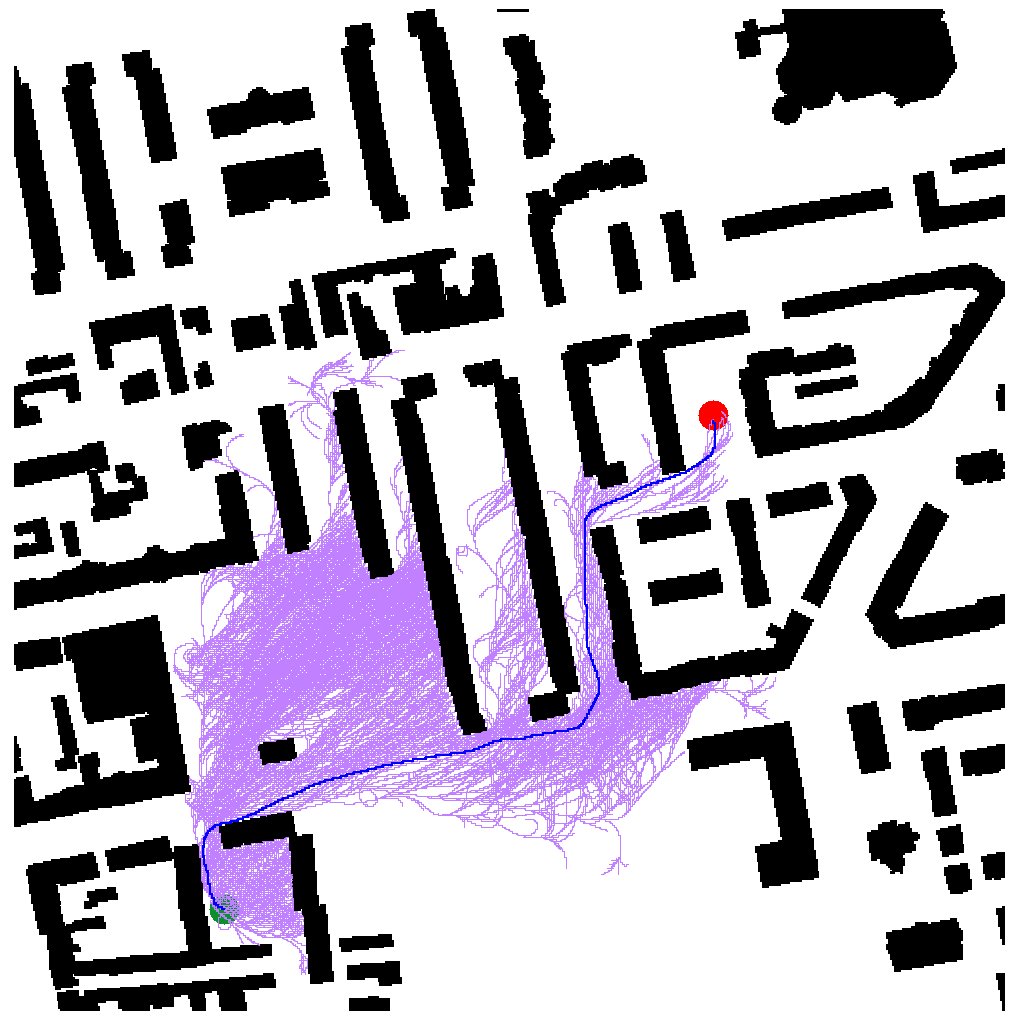}
    \end{subfigure} \\
    \begin{subfigure}{\linewidth}
    \includegraphics[width=0.49\linewidth]{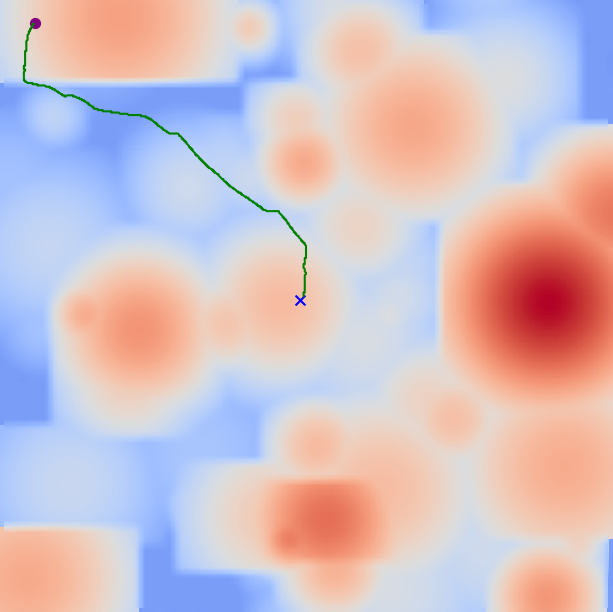}
    \includegraphics[width=0.49\linewidth]{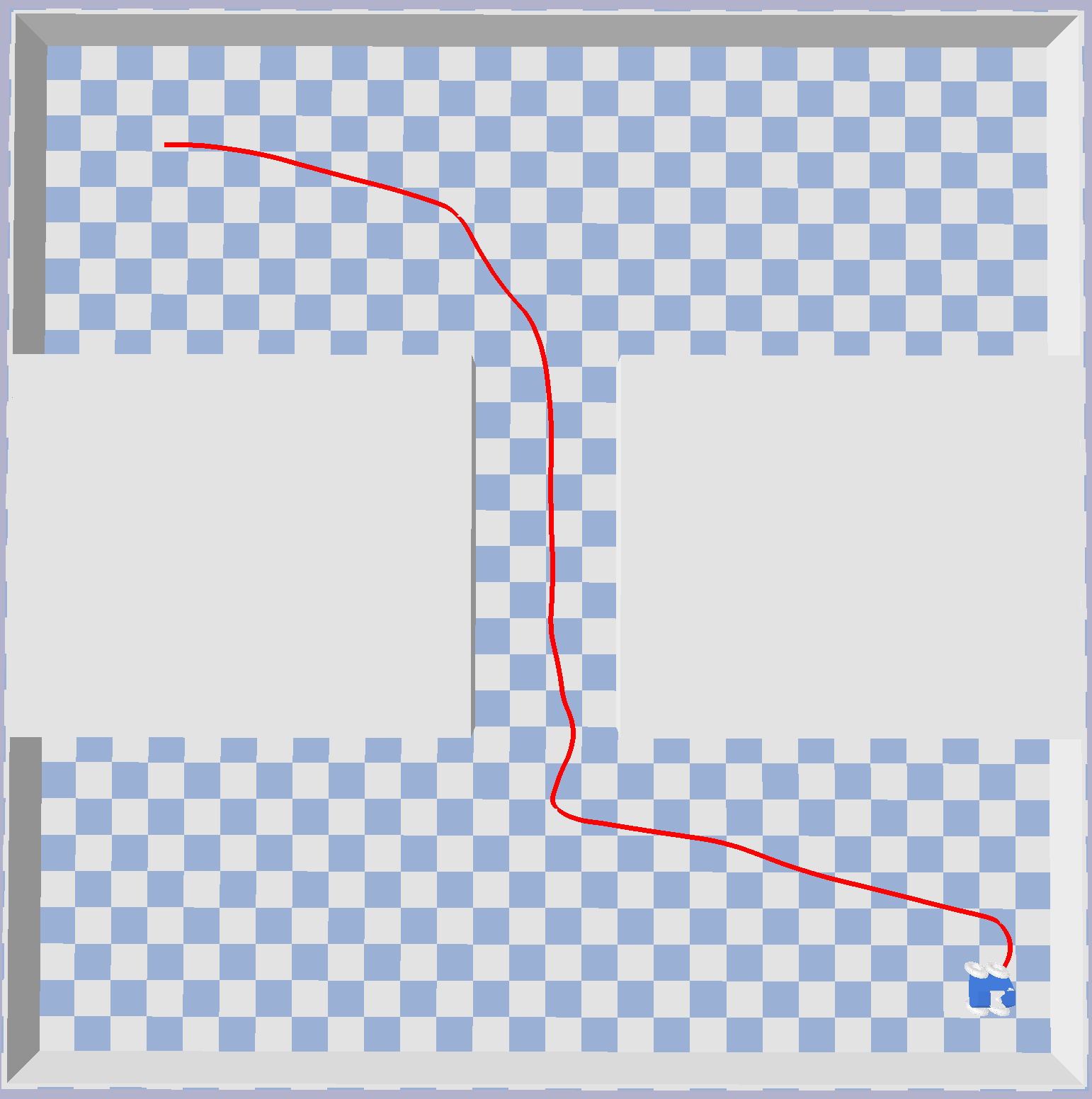}
    \end{subfigure} \\
    \caption{{\small Visualizations of solution trajectories found by the proposed method for the first-order (top-left, bottom-left over a cost-map) and second-order (top-right) differential drive systems, and a Segway robot (bottom-right) simulated using the Bullet physics engine.}}
    \vspace{-5mm}
    \label{fig:introduction}
\end{figure}

A representation of the planning environment, either in the form of binary occupancy grids, or cost maps that represent the cost of navigating across the environment, has been used as input to neural network models, which can then bias the planner towards shortest paths \cite{ichter2018learnedsampling,Kumar2019LEGOLE,wang2020neural,ichter2020criticalroadmaps}. Although it is possible to use these learned sampling strategies as input to a learned local planner, some of these samples may correspond to dynamically infeasible states or inevitable collision states (ICS). Recent work has proposed learning a sampling strategy for local goals that can be reached while obeying kinodynamic constraints and avoiding collisions \cite{johnson2020mpnet}, but assumes a steering function that can move the robot from one state to another.

This work proposes a learning process for a controller and integrates it with a framework for sampling-based kinodynamic planning to improve performance. The controller is trained offline using DRL in an empty environment to generate high quality controls towards desirable local goals. During planning, a node expansion procedure generates local goals biasing towards the global goal that are passed to the learned controller, which outputs the best control. The benefit of this two-step approach is that the learning process has to only deal with the system's dynamics and needs to be trained once per system. In this way it does not require a separate training process for each new environment that a robot is being deployed. The local goal selection and the planning framework are responsible for dealing with different problem instances and selecting a direction of progress towards the global goal. For the first expansion out of a node of the sampling-based tree, an informed approach is used where the local goal is selected so as to bias expansion towards the ultimate planning goal. Consecutive expansions follow a more exploratory strategy and select randomly local goals in the vicinity of the tree node. 

The evaluation uses a first-order and second-order differential drive system as well as a physically simulated Segway robot across different environments. The testing environments are different from the obstacle-free environment used for training the controller. It shows that the proposed framework results in very low cost solutions that are achieved faster relative to traditional random propagation approaches.

%% file: sections/02_problem.tex
\section{Problem Setup and Notation}
\label{sec:problem}

Consider a system with state space $\mathbb{X}$ - divided into  collision-free ($\mathbb{X}_f$) and obstacle ($\mathbb{X}_{obs}$) subsets - and control space $\mathbb{U}$, governed by the dynamics $\dot{x} = f(x,u)$ (where $x \in \mathbb{X}, u \in \mathbb{U}$).  The process $f$ can be an analytical ordinary differential equation (ODE) or modeled via a physics engine like Bullet \cite{coumans2019}. A \textit{plan} is defined by a sequence of piecewise-constant controls executed in order. A plan of length $T$ induces a trajectory $\tau \in \mathcal{T}$, where $\tau: [0,T] \mapsto \mathbb{X}_f$. For a start state $x_0 \in \mathbb{X}_f$ and a goal set $X_G \subset \mathbb{X}_f$, a feasible motion planning problem admits a solution trajectory of the form $\tau(0) = x_0, \tau(T) \in X_G$. Each solution trajectory has a cost $g$ according to a function $\texttt{cost}: \mathcal{T} \mapsto \mathbb{R}^+$, which the motion planner must minimize. In this setup, $X_G$ is defined as the set of all states that reach a specific state $x_G \in \mathbb{X}_f$ with some predefined tolerance.


%% file: sections/03_framework.tex
\section{Sampling-Based Motion Planning Framework}
\label{sec:framework}

We focus on sampling-based motion planners that only have access to a forward propagation model of the system's dynamics. This class of algorithms explore $\mathbb{X}_f$ by incrementally constructing a tree by sampling $\mathbb{U}$. Algorithm~\ref{alg:tree-sbmp} outlines the high-level operation of a sampling-based motion planner (Tree-SBMP) that builds a tree of states reachable from $x_0$. 

Each iteration of Tree-SBMP starts with the selection of an existing tree node to expand that corresponds to a robot state (Line 3). A candidate control is considered for the selected node, and an end state is obtained by forward propagation of the system dynamics (Lines 4-5). The resulting edge is added to the tree if it is not in collision (Lines 6-7). By varying how these operations are implemented, different algorithms can be obtained. After $N$ iterations (or a preset time limit), if the tree has states in $X_G$, the best found solution trajectory according to $g$ can be returned.

\vspace{-.05in}
\begin{algorithm}[h!]
\SetAlgoLined
$T \leftarrow \{x_0\}$; \\
\While{termination condition is not met}
{
$x_{select} \leftarrow \texttt{SELECT-NODE}(T)$; \\ 
$u \leftarrow \texttt{EXPAND-NODE}(x_{select})$; \\
$x_{new} \leftarrow \texttt{PROPAGATE}(x_{select},u)$; \\
\If{$(x_{select} \rightarrow x_{new}) \in \mathbb{X}_f$}
{
\texttt{EXTEND-TREE}($T, x_{select} \rightarrow x_{new}$);
}
}
\caption{Tree-SBMP($\mathbb{X}_f,\mathbb{X}_o,\mathbb{U},x_0,X_G$)}
\label{alg:tree-sbmp}
\end{algorithm}
\vspace{-.05in}

\begin{figure}[h!]
    \centering
    \includegraphics[width=0.475\textwidth]{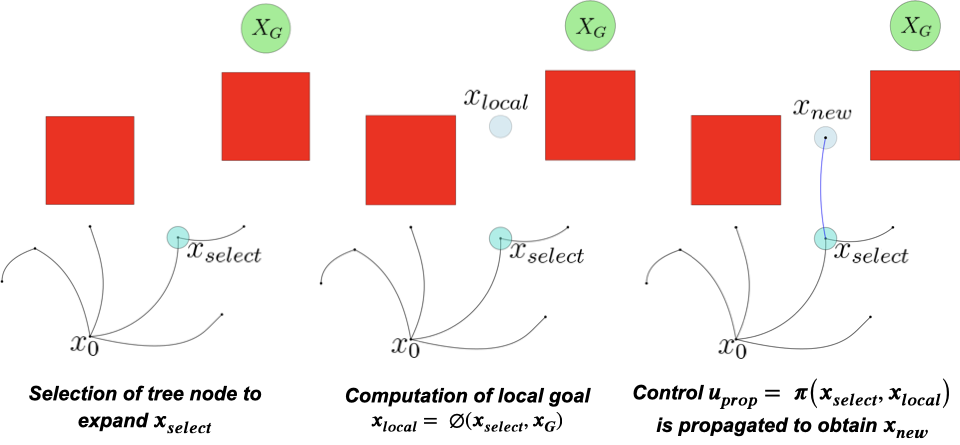}
    \caption{Illustration of the proposed expansion function. }
    \label{fig:method-1}
    \vspace{-2.5mm}
\end{figure}

This work presents a version of \texttt{EXPAND-NODE} (Fig.~\ref{fig:method-1}) that selects a local goal in the environment and uses a learned controller to predict a control that makes progress to the local goal. The proposed node expansion framework is integrated into the DIRT algorithm \cite{LB-DIRT}, which also achieves AO properties. DIRT is selected due to its informed implementation of \texttt{SELECT-NODE} that biases selection towards high quality nodes close to the unexplored parts of the state space.

\vspace{-.05in}

DIRT also evaluates multiple candidate controls as part of \texttt{EXPAND-NODE}. This \textit{Blossom expansion} procedure ranks the resulting trajectories and processes them in order so as to prioritize expansions that bring the system closer to the goal according to a heuristic cost-to-go. 



%% file: sections/04_method.tex
\section{Proposed Method}
\label{sec:method}

The proposed pipeline has two stages. \textit{Offline:} For each system, a controller is trained using DRL to reach a goal state in an obstacle-free environment. For every planning environment, a representation in the form of an obstacle map or of a cost map is assumed available.  \textit{Online: } Upon planner initialization, a vector field is constructed for the given planning problem to bias the planner. At every iteration, a local goal state is generated as input to the learned controller. The controller then outputs a candidate control to be propagated towards the local goal state. The local goal is generated either in an exploitative manner using the goal-biasing vector field or in an exploratory manner by using random sampling.

\subsection{Learning the controller}

Given the current state of the system $x \in \mathbb{X}$, the aim is to train a controller $u_{RLC} = \pi(x,x_G)$ to reach the goal set $X_G$ in the absence of obstacles. Let $\rho_\pi$ be the distribution of trajectories induced by the controller $\pi$, and $x_0$ the initial state sampled from an initial state distribution $\rho(x_0)$. The controller can be optimized using a DRL algorithm that maximizes the \textit{expected return} \vspace{-0.05in}$$\mathcal{L}(\pi) = \mathbb{E}_{\tau \sim \rho_\pi} [\sum_{t=0}^\infty \gamma^t r(x_t, x_G)],\vspace{-0.05in}$$ where  $\gamma \in [0,1]$ is a discount factor and $r: \mathbb{X} \times \mathbb{X} \mapsto \{-1,0\}$ is a bounded reward function defined as $r(x_t, x_G) = 0$ if $x_t \in X_G$ and $r(x_t, x_G) = -1$ otherwise. Alternative DRL algorithms can be used to maximize $\mathcal{L}(\pi)$. This work employs Hindsight Experience Replay (HER) \cite{Andrychowicz2017HindsightER} to facilitate learning from sparse rewards.

\vspace{-0.05in}
\begin{figure}[h!]
    \centering
    \includegraphics[width=\linewidth]{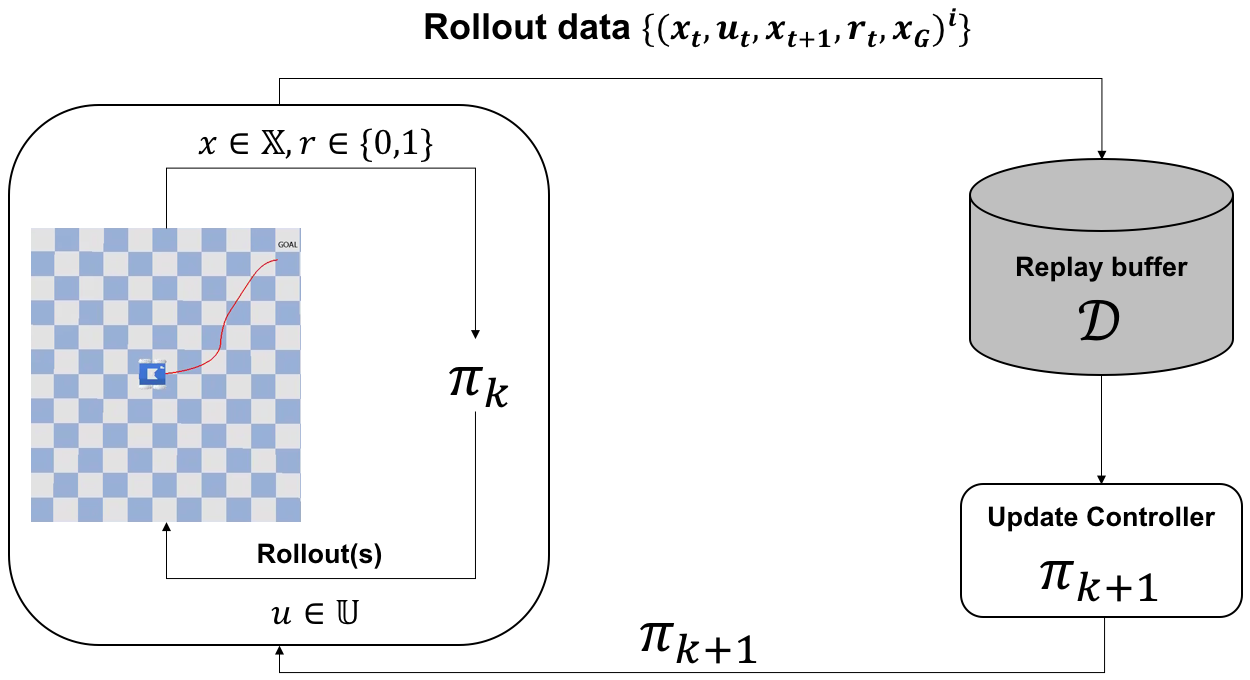}
    \caption{{\small Off-policy DRL algorithm with online data collection. The replay buffer $\mathcal{D}$ is implemented using Hindsight Experience Replay (HER). The controller update step depends on the choice of the DRL algorithm used to maximize the expected return.}}
    \vspace{-2.5mm}
    \label{fig:method-3}
\end{figure}

As the focus is mainly on vehicular systems, $x_G$ is defined relative to the robot's current state. For the state variables of the system that are transformation-invariant (e.g., current position and steering angle), the initial state of the system is fixed at the origin. The values of the initial state variables that are not transformation-invariant (e.g., velocities of the wheels) are sampled uniformly within bounds. This re-parameterization reduces data requirements while learning the controller.

{\bf Comparison alternative:} A Supervised Learned Controller (SLC), $u_{SLC} = \pi(\delta x)$ is also evaluated. It learns the inverse of a discretized set of forward dynamics. SLC acts as a locally optimal surrogate for the system's dynamics. The control-duration space is discretized to form a discrete set of plans $\mathcal{U}_{LC}$ and all plans in $\mathcal{U}_{LC}$ are forward propagated from a fixed initial state. The resulting end states form a reachability region $\mathcal{X}$. A dataset is obtained by using a nearest neighbor data structure to store the best plan that reaches a neighborhood in $\mathcal{X}$ for every initial state. A neural network is then trained to minimize the mean squared error (MSE) between the plan in the dataset and the plan output by SLC given $\delta x$, which is defined as the \textit{difference in states} between an initial state $x_0 \in \mathbb{X}$ and a local goal state $x_G \in \mathcal{X}$. 


\subsection{Generating informed local goal states among obstacles}


A two-dim. workspace $\mathbb{T}$ can be divided into two disjoint subsets: the collision-free subset $\mathbb{T}_f$ and the obstacle subset $\mathbb{T}_o$ containing all $n$ obstacles. The boundary of the environment is also considered as an obstacle. Define points $p$ as projections of states $x$ from $\mathbb{X}$ to $\mathbb{T}$. Then, let $p_f \in \mathbb{T}_f$ be the projection in $\mathbb{T}$ of each tree state/node $x_{select}$ selected for propagation. Then, the objective is to generate a local goal point $p_{lg} \in \mathbb{T}_f$ so as to perform the propagation with the learned controller. Given the global goal $p_G$, $p_{lg}$ is generated by a \textit{local goal function} $\phi(p_f,p_G)$ so that: i) $h^*(p_{lg}) < h^*(p_f)$ where $h^*$ is a cost-to-go function given $p_G$; and ii) $p_{lg}$ maximizes clearance from obstacles.


\begin{wrapfigure}{r}{0.25\textwidth}
\vspace{-.1in}
    \centering
    \includegraphics[width=0.24\textwidth]{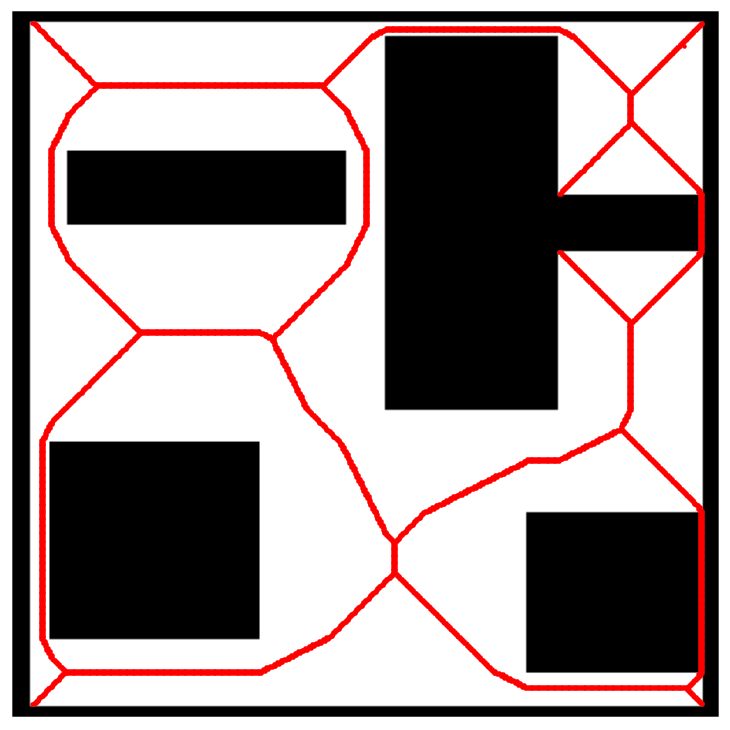}
    \vspace{-.1in}
    \caption{{\small An example $\mathcal{MA}$ highlighted in red.}}
    \vspace{-.2in}
    \label{fig:medial-axis}
\end{wrapfigure}
The local goal function $\phi(p_f,p_G)$ is implemented using a \textit{medial axis} $\mathcal{MA} \subset \mathbb{T}_f$. Given an obstacle surface $\mathcal{O}_i \subset \mathbb{T}_o$ (e.g., an edge of a polygonal obstacle in 2D) there exists a subset $P_{\mathcal{O}_i} \subset \mathbb{T}_f$ of all free points, such that $\mathcal{O}_i$ is the closest obstacle surface. The medial axis is the intersection of all such subsets: $\mathcal{MA} = \bigcap_{i=0}^n  P_{\mathcal{O}_i}$.  To compute the medial axis, the environment is discretized. The computation is performed once for every planning environment (Fig~\ref{fig:medial-axis}). 





\begin{figure}[h!]
    \centering
    \begin{subfigure}{\linewidth}
    \includegraphics[width=0.49\linewidth]{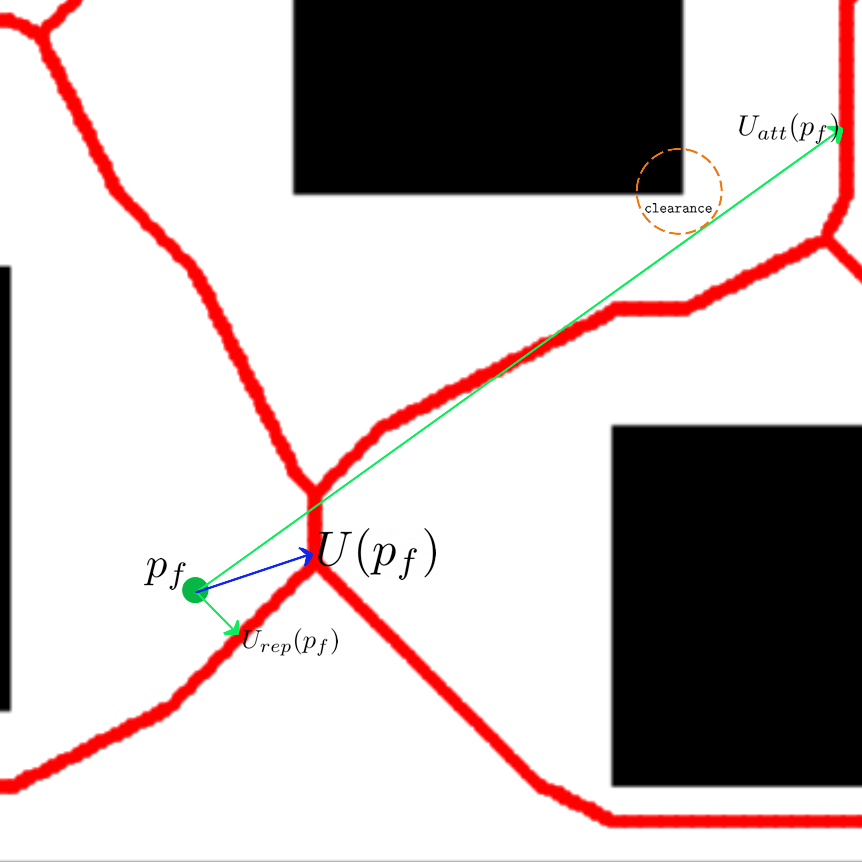}
    \includegraphics[width=0.49\linewidth]{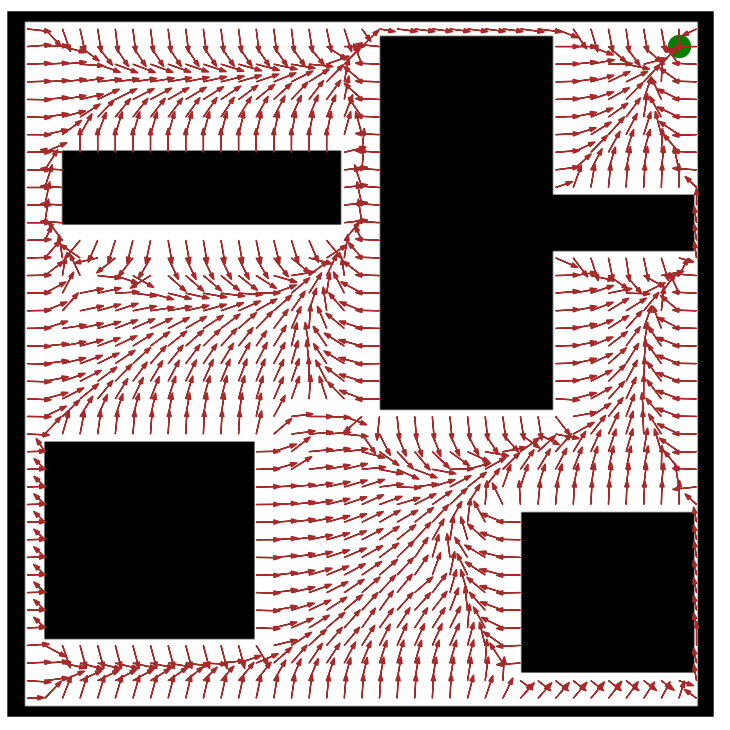}
    \end{subfigure}
    \caption{{\small(Left) Closeup of $\mathcal{MA}$ from Fig. \ref{fig:medial-axis}. Given $p_f \in \mathbb{T}_f$, the vectors $U_{rep}(p_f)$ and $U_{att}(p_f)$ give rise to $U(p_f)$. (Right) The normalized vector field for a given goal (green circle). }}
    \vspace{-2.5mm}
    \label{fig:medial-axis_2}
\end{figure}

For a given $p_f \in \mathbb{T}_f$, 3 vectors are computed (Fig~\ref{fig:medial-axis_2} left). First, a repulsive vector $U_{rep}(p_f)$ that points $p_f$ away from the closest obstacle towards the nearest point on $\mathcal{MA}$. An attractive vector $U_{att}(p_f)$ that points to $p_{MA}^* \in (\mathcal{MA} \cup p_G)$ such that: (a) $p_f$ has line of sight (with clearance) to $p_{MA}^*$, and (b) $h^*(p_{MA}^*)$ is minimum, i.e., makes the most progress to the goal among all visible points from $p_f$. Both vectors are used to compute $U(p_f) = w \cdot U_{rep}(p_f) + (1 - w) \cdot U_{att}(p_f)$, where $w = \vert U_{rep}(p_f) \vert / (\vert U_{rep}(p_f) \vert + \vert U_{att}(p_f) \vert))$. The magnitude of $U$ is set so it \textit{reaches} $\mathcal{MA}$ when applied at $p_f$. $U$ is the final output of $\phi(p_f,p_G)$. The normalized version of $U$ is shown in  Fig~\ref{fig:medial-axis_2} right.  It is computed once offline given a discretization of the environment.

\subsection{Generating informed local goal states given a costmap}

\begin{figure}[h!]
    \centering
    \begin{subfigure}{\linewidth}
    \includegraphics[height=0.165\textheight]{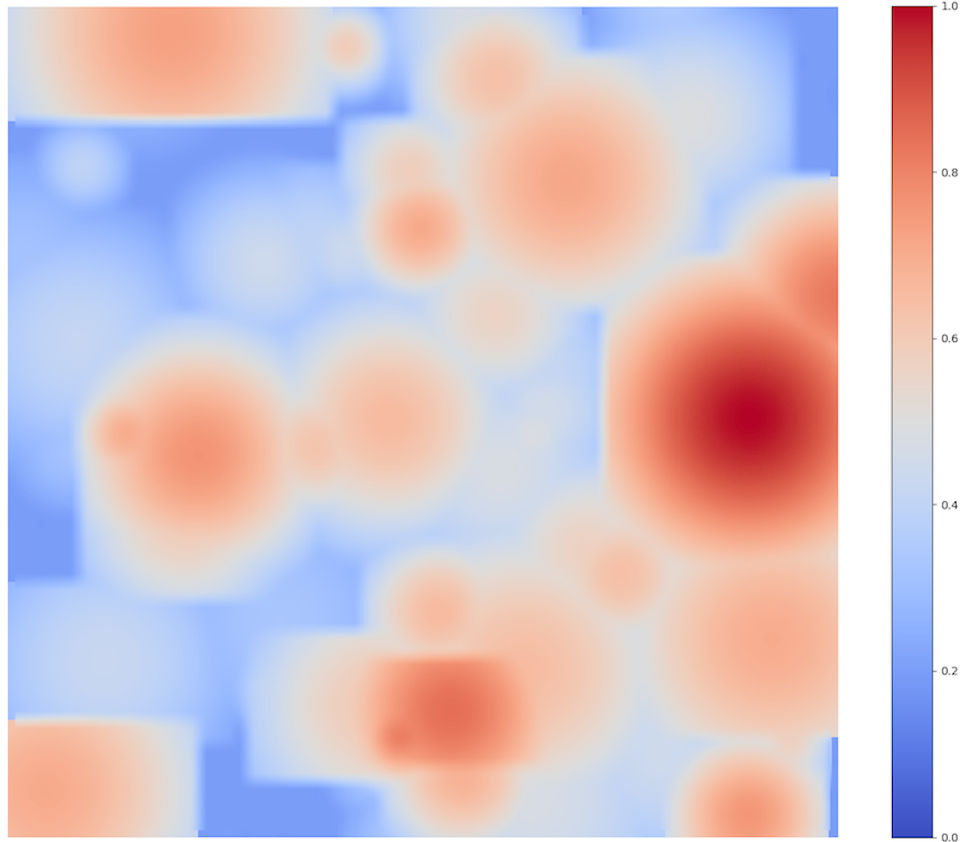}
    \includegraphics[height=0.165\textheight]{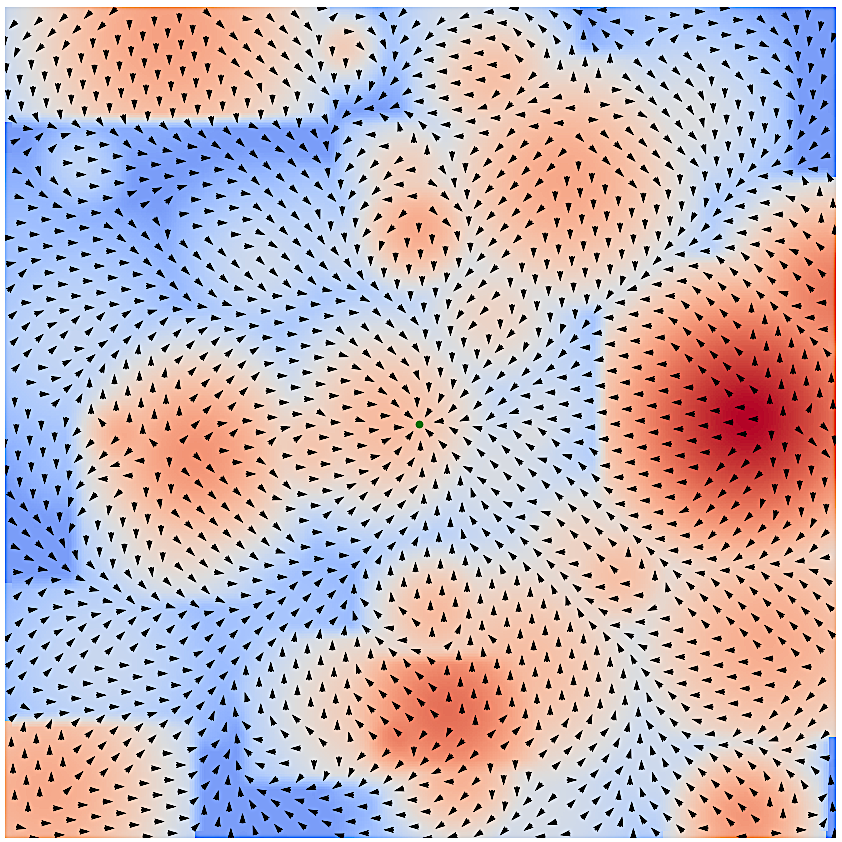}
    \end{subfigure}
    \vspace{-.05in}
    \caption{{\small (Left) Cost map of a planning environment where blue/red regions are easy/hard to traverse. (Right) The $U$ vector field for a given goal state (green circle).}}
    \vspace{-2.5mm}
    \label{fig:wavefront}
\end{figure}

To use the learned controller $\pi$ in the traversability cost environment, the environment is discretized on a grid in $\mathbb{T}$. For every $p \in \mathbb{T},$ the traversability cost $c_{tv}(p) \in [0,1]$ is defined, where $c_{tv}(p) < 1$ is traversable (Fig.~\ref{fig:wavefront}, left). The higher the cost, the longer it takes for the robot to traverse. Equation \eqref{eq:cmap_cost} computes the path cost of a trajectory $\tau$ of length $T$ as: \vspace{-.2in}

\begin{equation} \label{eq:cmap_cost}
    \texttt{cost}(\tau) = \int_0^T \exp(K c_{tv}(\tau(t))) dt 
    \vspace{-.2in}
\end{equation}

Offline, once per planning problem, a modified version of the any-angle path planning algorithm $\text{Theta}^*$ \cite{thetastar} is computed over the grid to create a vector field (Fig.~\ref{fig:wavefront}, right). Instead of checking for an obstacle-free line of sight when smoothing the solution path, the modified version checks for a line of decreasing cost of a minimum length. For a point $p \in \mathbb{T}$, the vector $U(p)$ points to the next node $p^*$ along the lowest cost path to the goal $p_G$ according to $\text{Theta}^*$. Online, the end point of $U$ is returned as the output of $\phi(p,p_G).$


%% file: sections/05_results.tex
\section{Results}
\label{sec:results}

The systems considered in our evaluation are an analytically simulated first-order ($\texttt{dim}(\mathbb{X}) = 3, \texttt{dim}(\mathbb{U}) = 2$) and second-order differential drive vehicles ($\texttt{dim}(\mathbb{X}) = 5, \texttt{dim}(\mathbb{U}) = 2$) and a Bullet-simulated Segway robot ($\texttt{dim}(\mathbb{X}) = 3, \texttt{dim}(\mathbb{U}) = 2$).

\subsection{Controller training evaluation}

The following algorithms were tested to learn $u_{RLC} = \pi(x,x_G)$: Deep Q-Networks (DQN) \cite{mnih2015human}, Twin-Delayed Deep Deterministic Policy Gradient (TD3) \cite{fujimoto2018addressing} and Soft Actor-Critic (SAC) \cite{Haarnoja2018SoftAO}. For DQN, the discretized action space $\mathcal{U}$ consists of $3^{\texttt{dim}(\mathbb{U})}$ controls. For each dimension of the control space, the control can take either the minimum, mid or maximum value. The training pipeline was implemented on a Google Colab CPU instance. The DRL algorithm implementations were derived from Gym \cite{openaigym} and Stable Baselines 3 \cite{stable-baselines3}.

\begin{figure}[h!]
    \centering
    \begin{subfigure}{\linewidth}
    \includegraphics[width=0.49\linewidth]{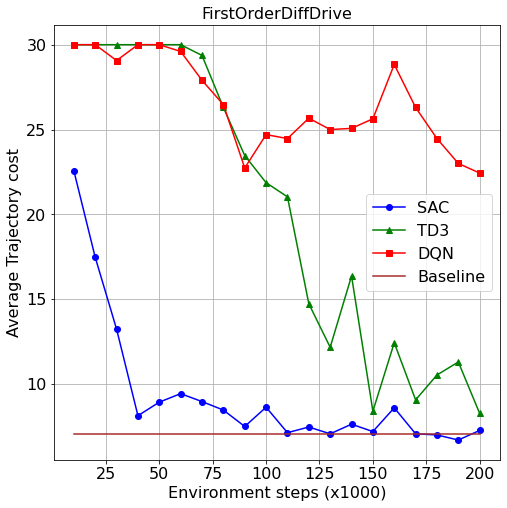}
    \includegraphics[width=0.49\linewidth]{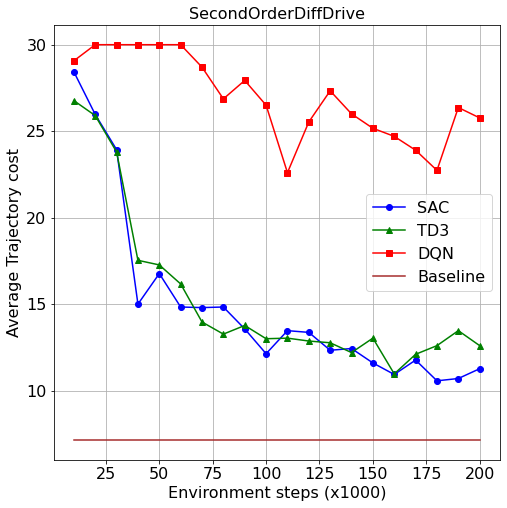}
    \end{subfigure}
    \caption{{\small Training performance of different controllers for the first-order and second-order differential drive systems. The baseline is chosen to be the highest possible cost that a goal-reaching trajectory can incur in an obstacle-free environment.}}
    \vspace{-7.5mm}
    \label{fig:training_results}
\end{figure}

Each controller is trained for 200 thousand forward propagation steps, and their performance is evaluated using the average time to reach the goal (trajectory cost). Fig.~\ref{fig:training_results} shows the results for the analytical systems. The costs are averaged over 30 trials, where each trial corresponds to 100 randomly sampled goals. DQN does not converge to a controller that finds low cost goal-reaching trajectories due to the restrictive nature of the control space. Although SAC and TD3 exhibit nearly similar performance after parameter tuning, the best-performing SAC controller is selected for the proposed expansion framework since its performance is closest to the baseline. Subsequently, SAC+HER  is also used to train the controller for the Segway system.

\begin{figure*}[ht!]
\centering
\begin{subfigure}[b]{\textwidth}
         \includegraphics[width=.245\textwidth]{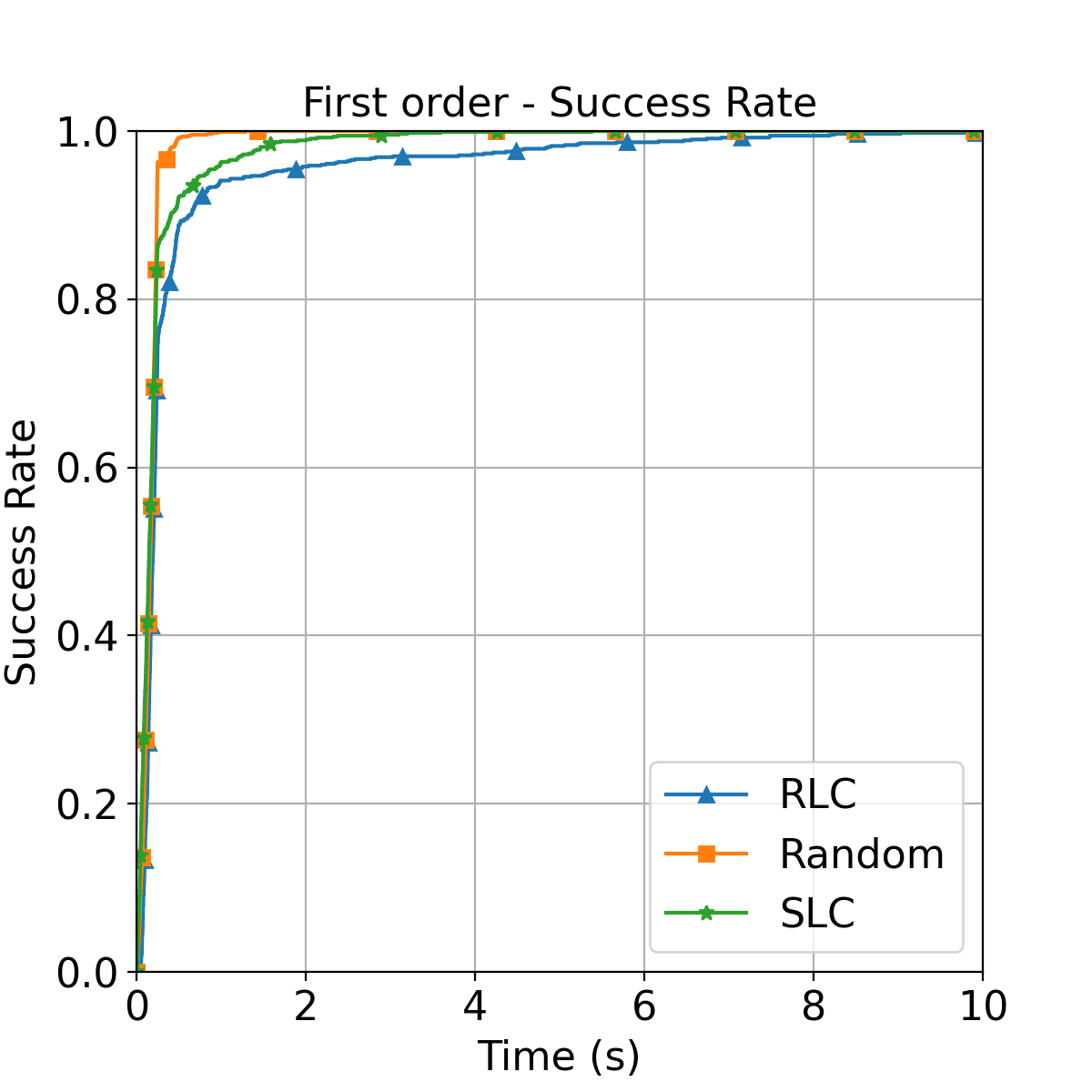}
         \includegraphics[width=.245\textwidth]{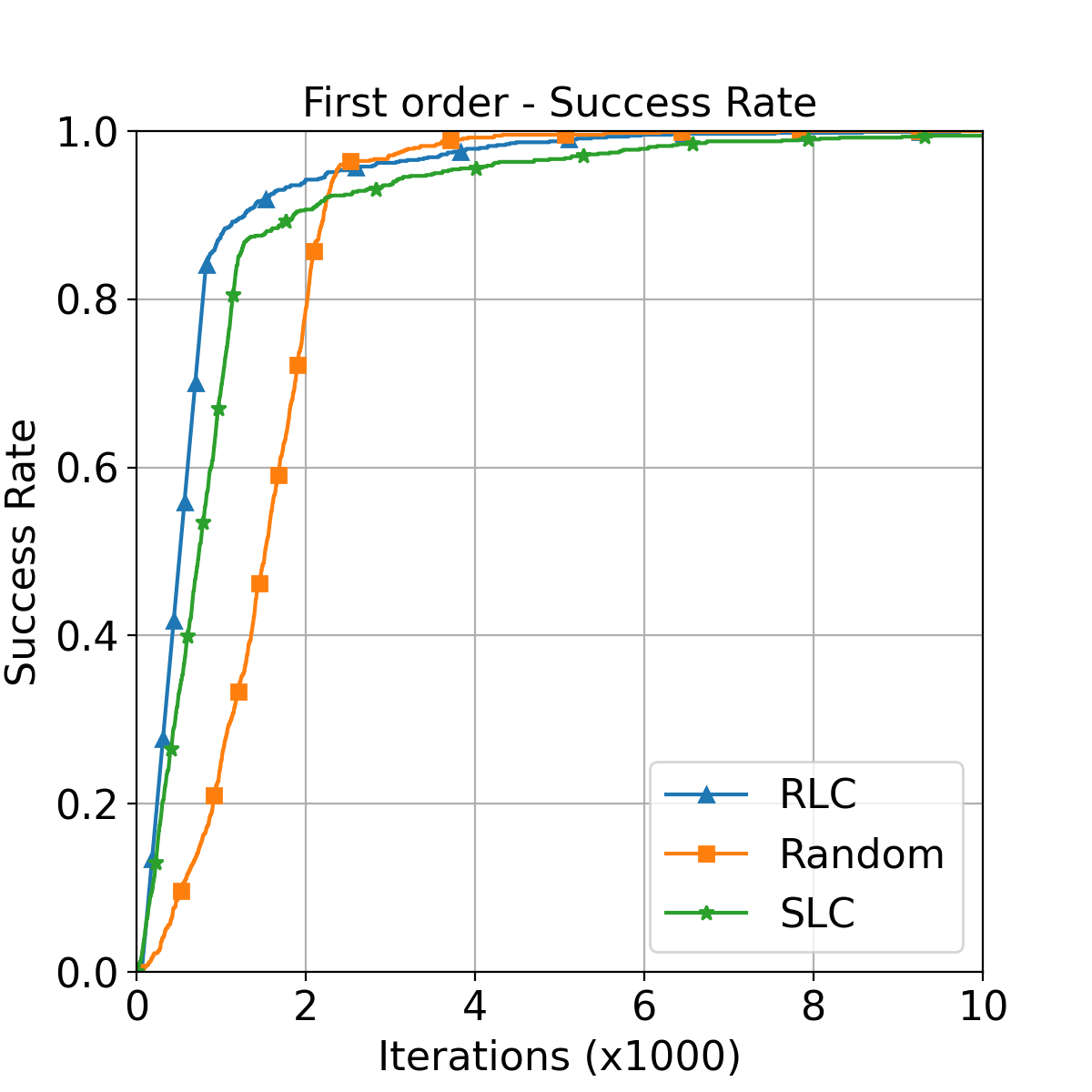}
         \includegraphics[width=.245\textwidth]{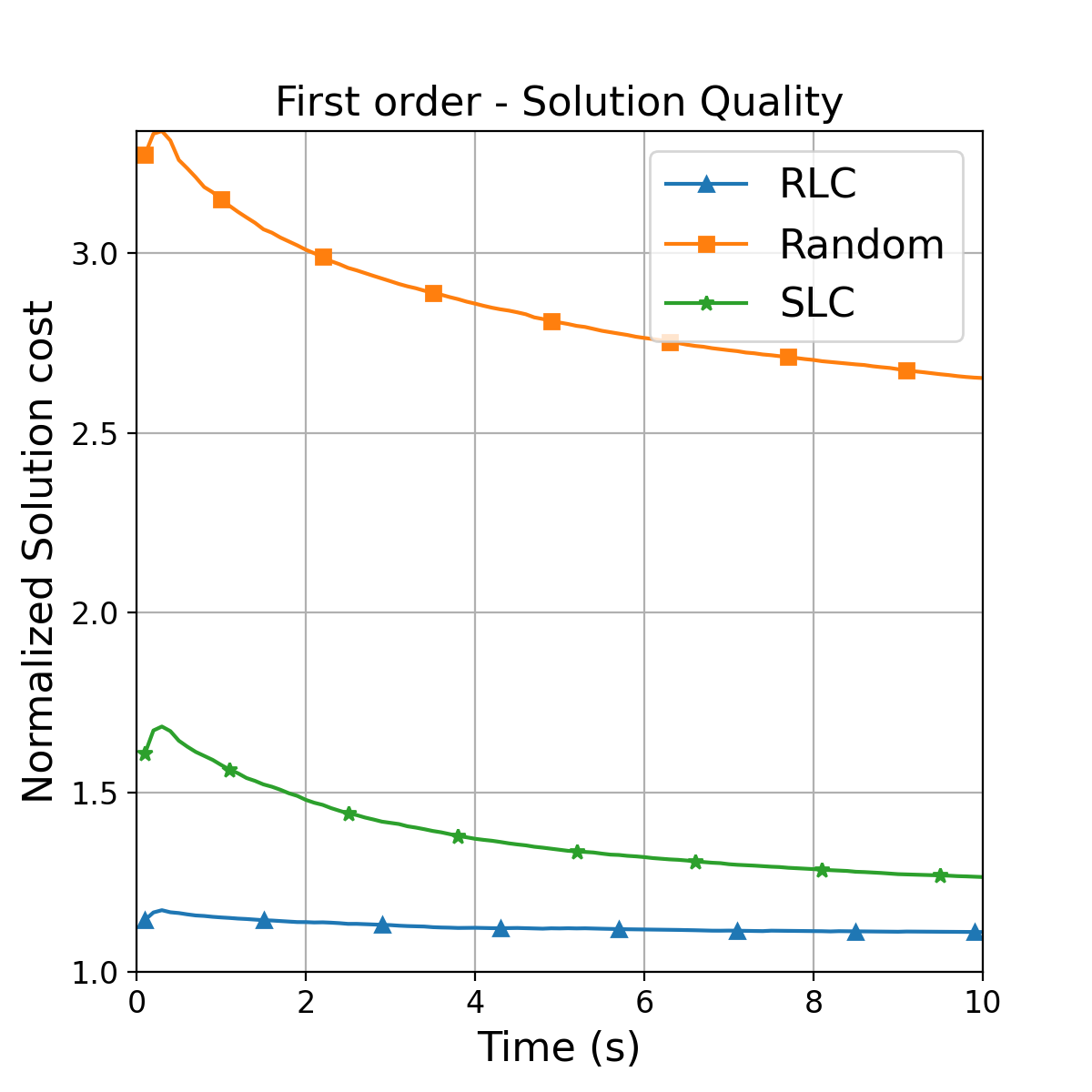}
         \includegraphics[width=.245\textwidth]{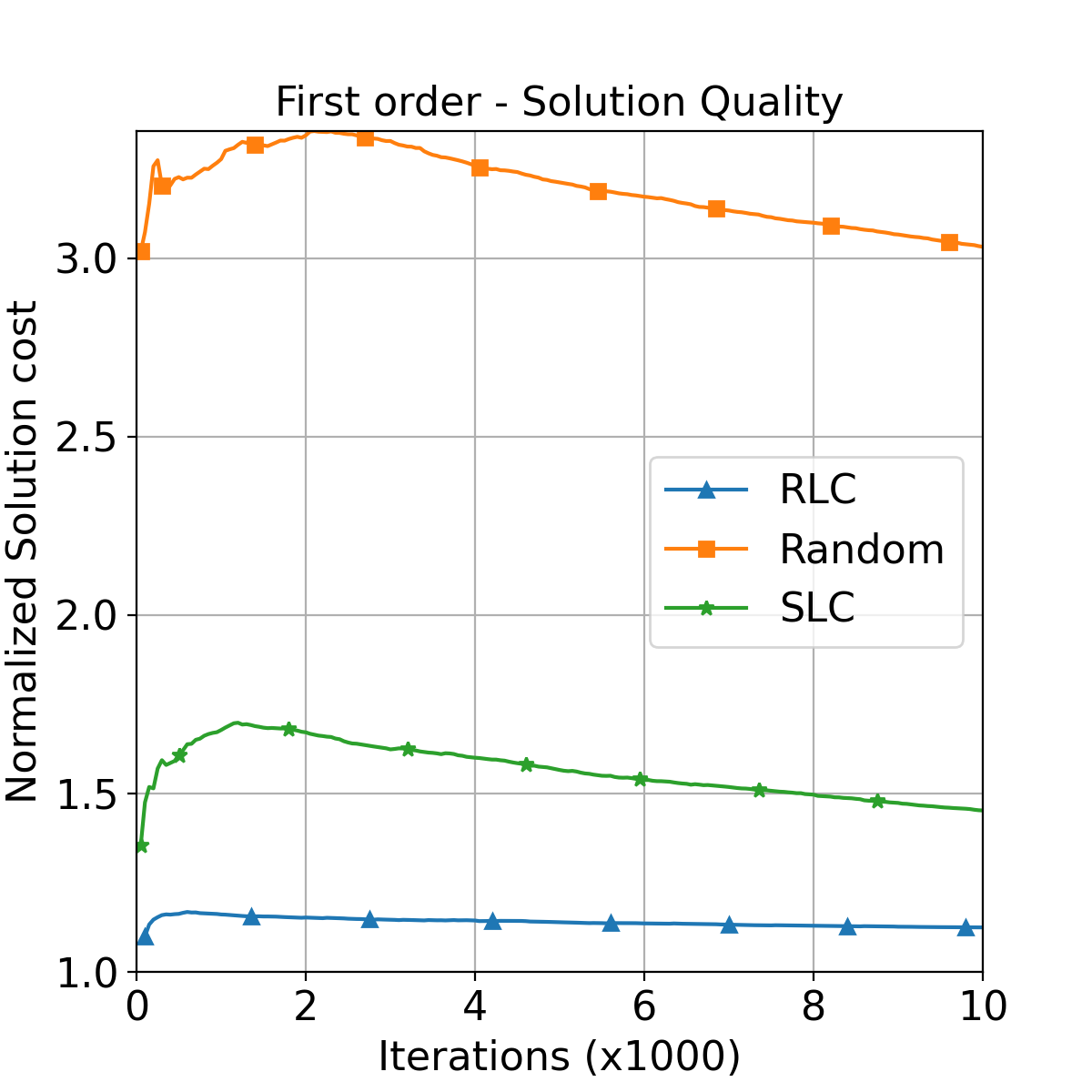}
         \vspace{-.2in}
         \caption{Results for first-order differential drive system in the city environments.}
         \label{fig:planner_analysis_fo}
\end{subfigure}\\
\begin{subfigure}[b]{\textwidth}
         \includegraphics[width=.245\textwidth]{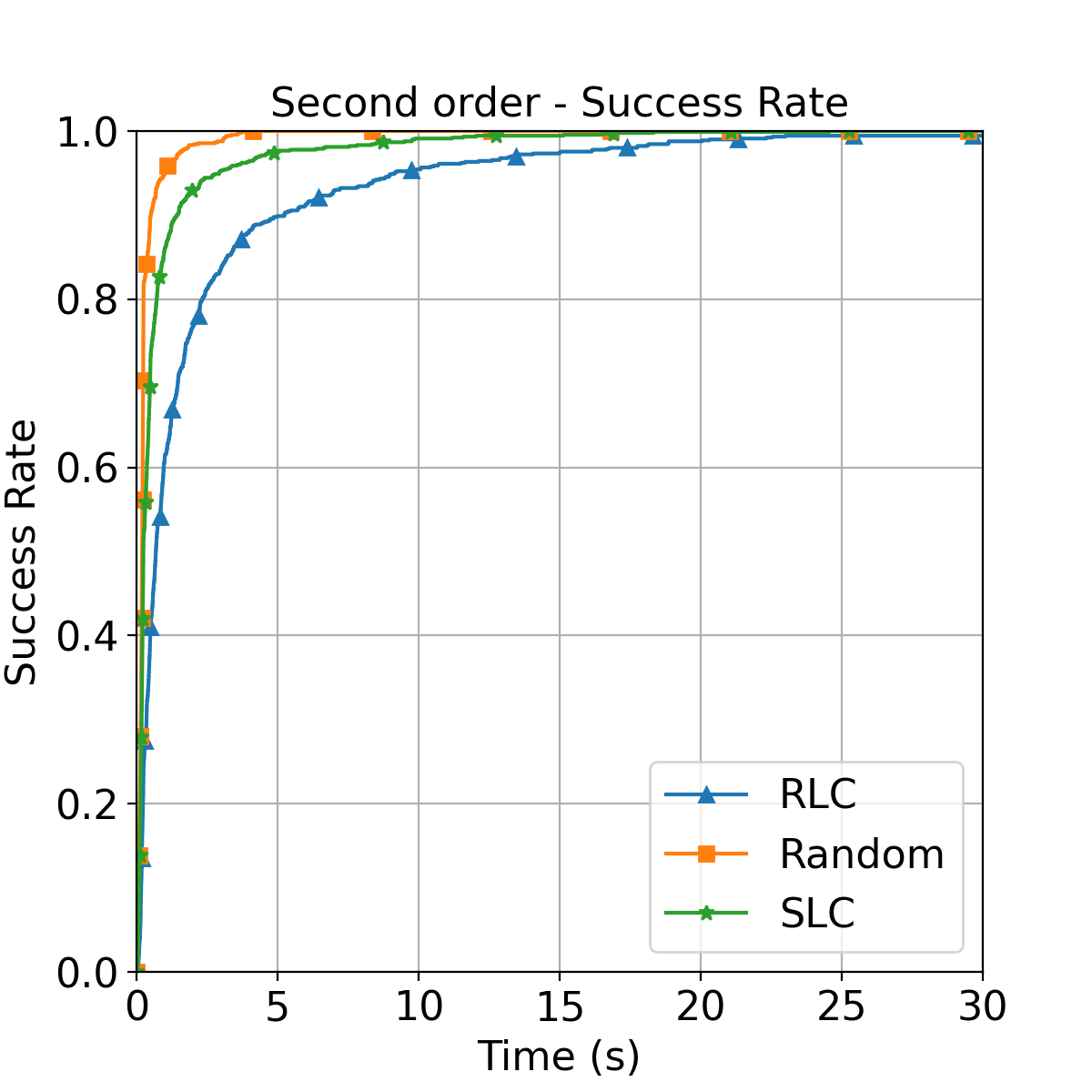}
         \includegraphics[width=.245\textwidth]{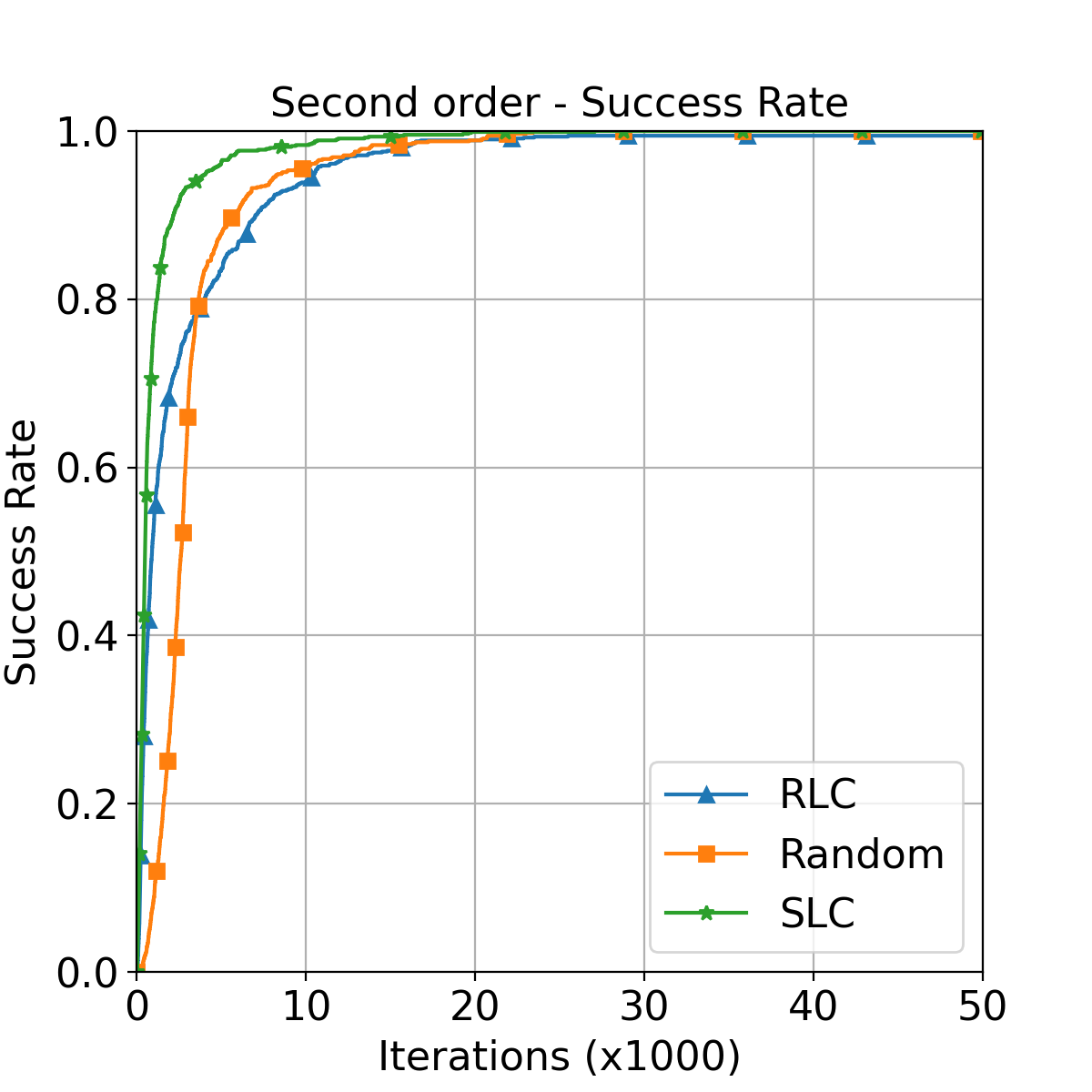}
         \includegraphics[width=.245\textwidth]{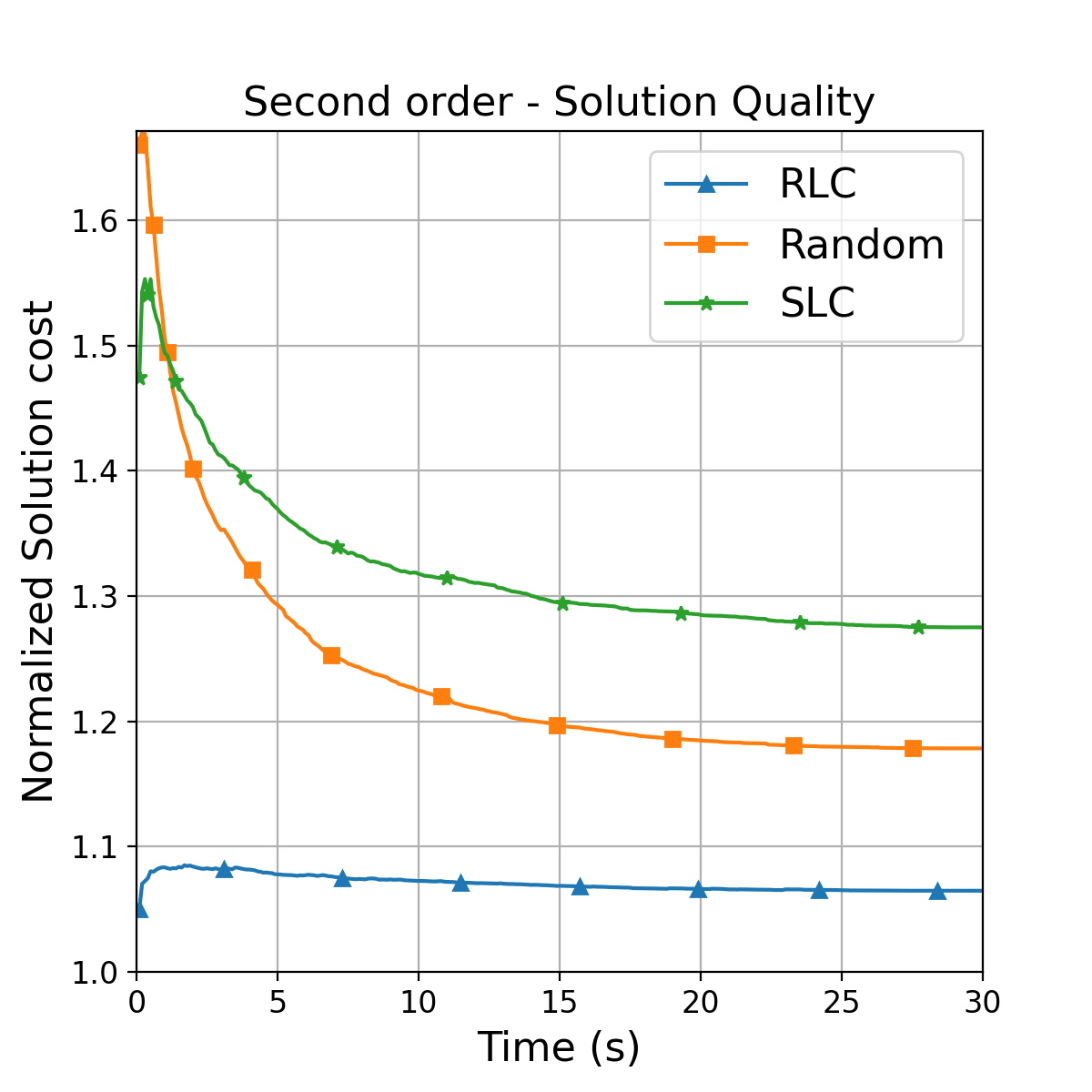}
         \includegraphics[width=.245\textwidth]{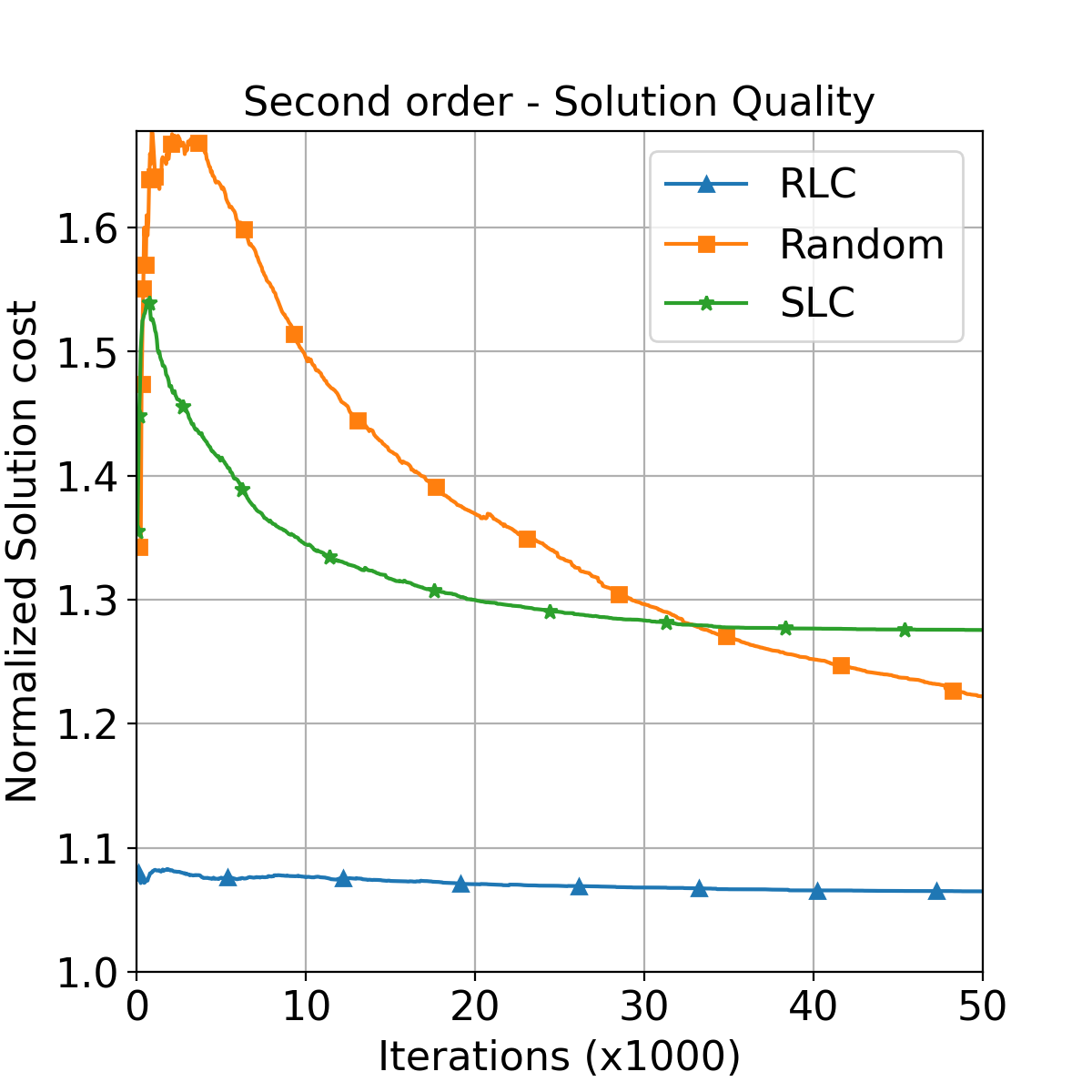}
         \vspace{-.2in}
         \caption{Results for second-order differential drive system in the city environments.}
         \label{fig:planner_analysis_so}
\end{subfigure}\\
\begin{subfigure}[b]{\textwidth}
         \includegraphics[width=.245\textwidth]{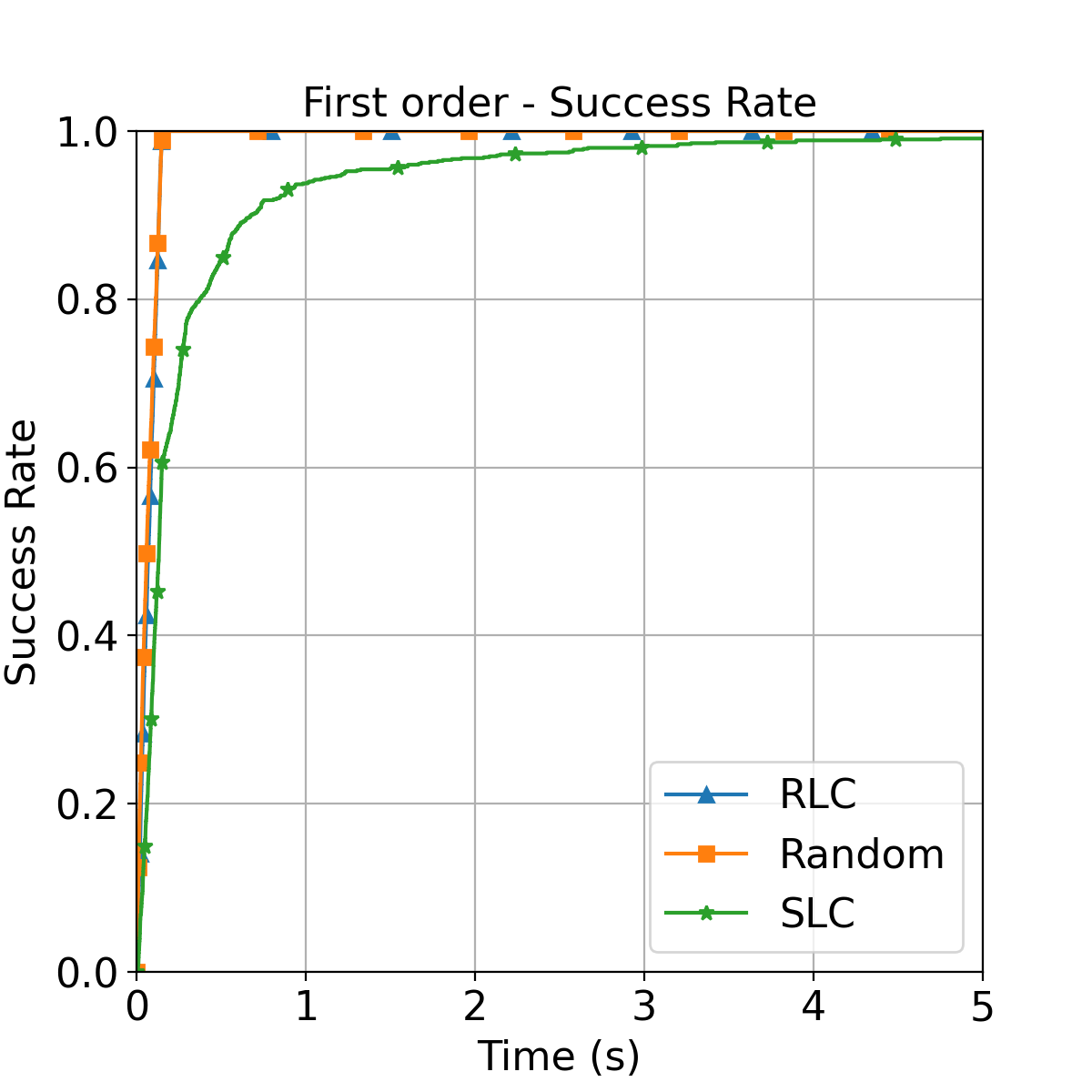}
         \includegraphics[width=.245\textwidth]{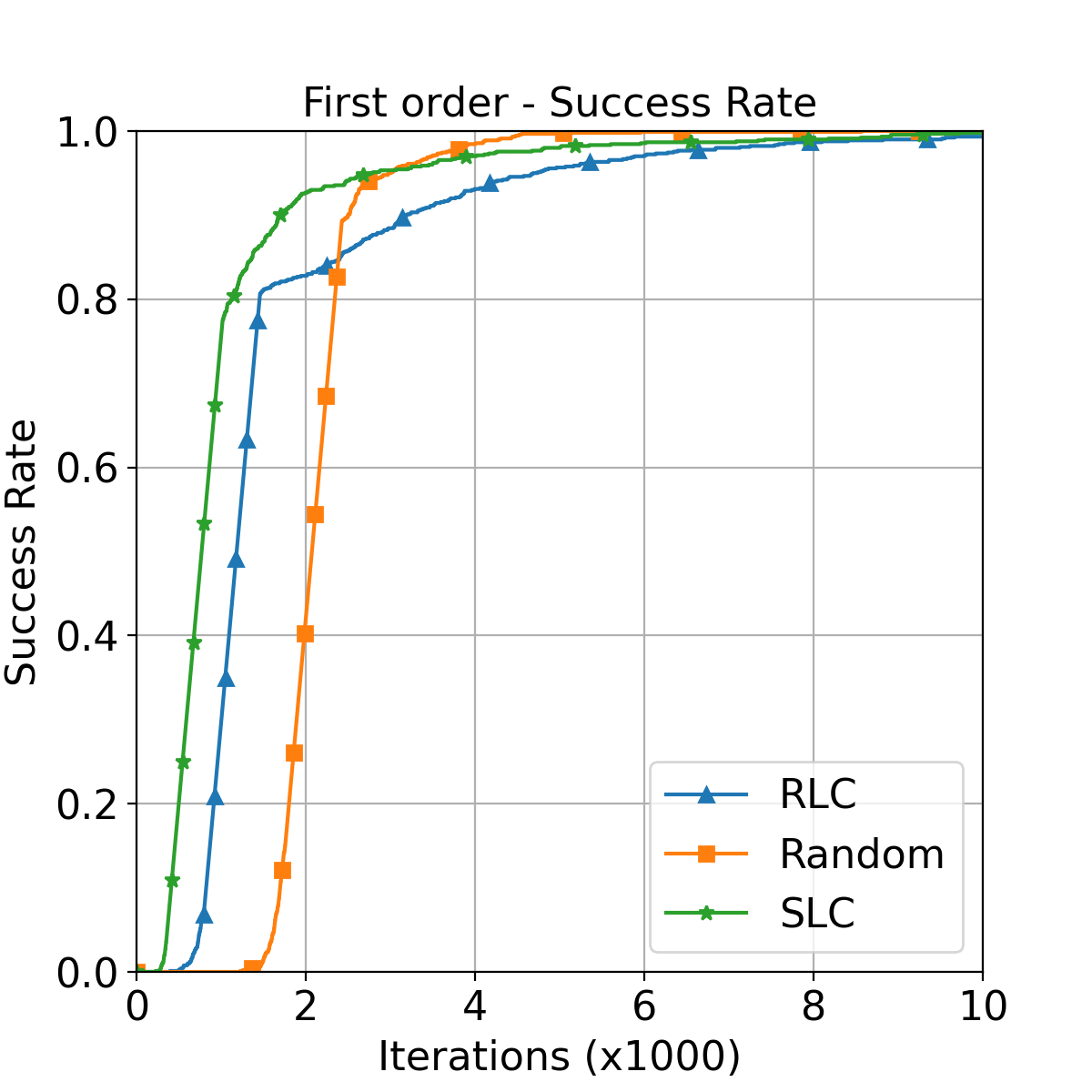}
         \includegraphics[width=.245\textwidth]{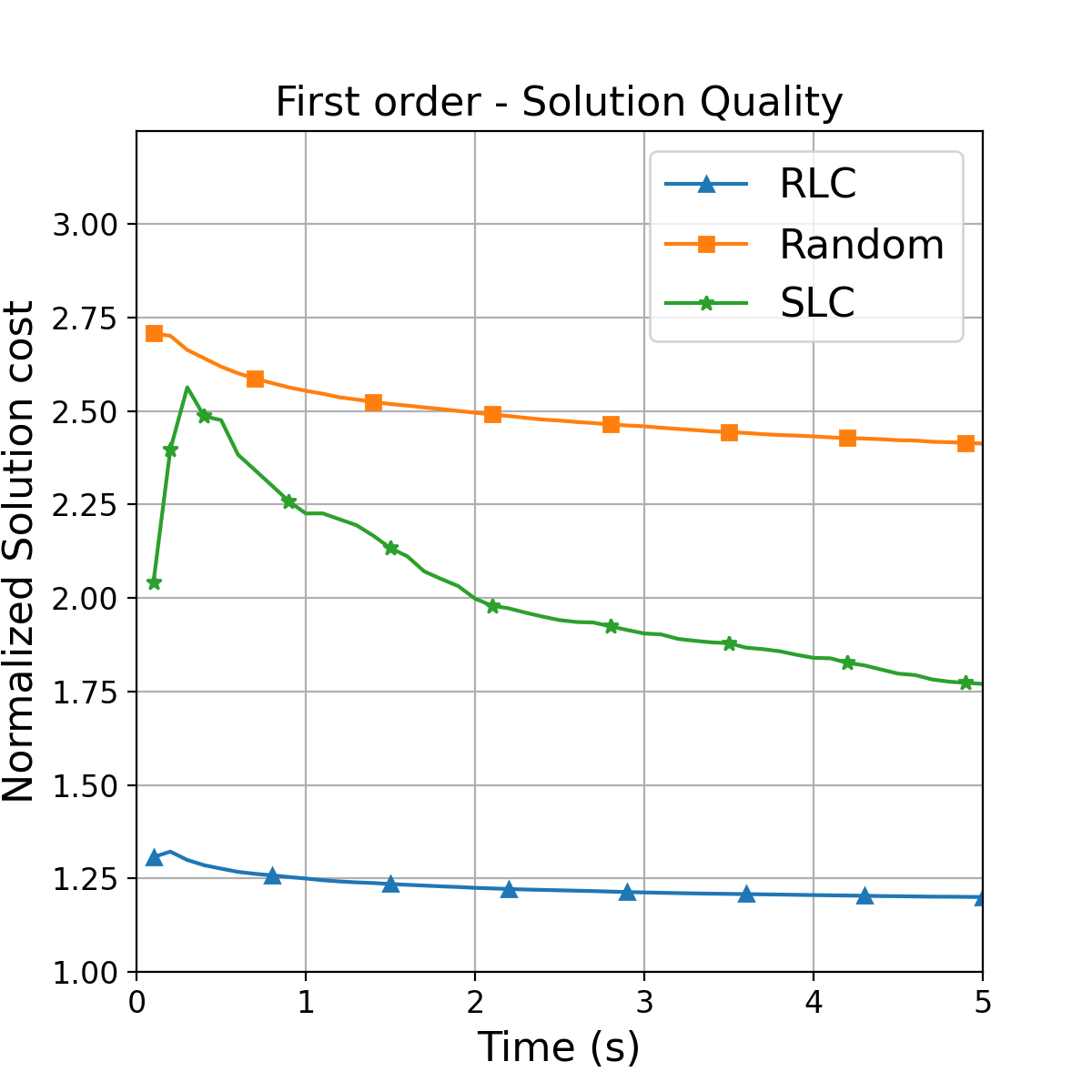}
         \includegraphics[width=.245\textwidth]{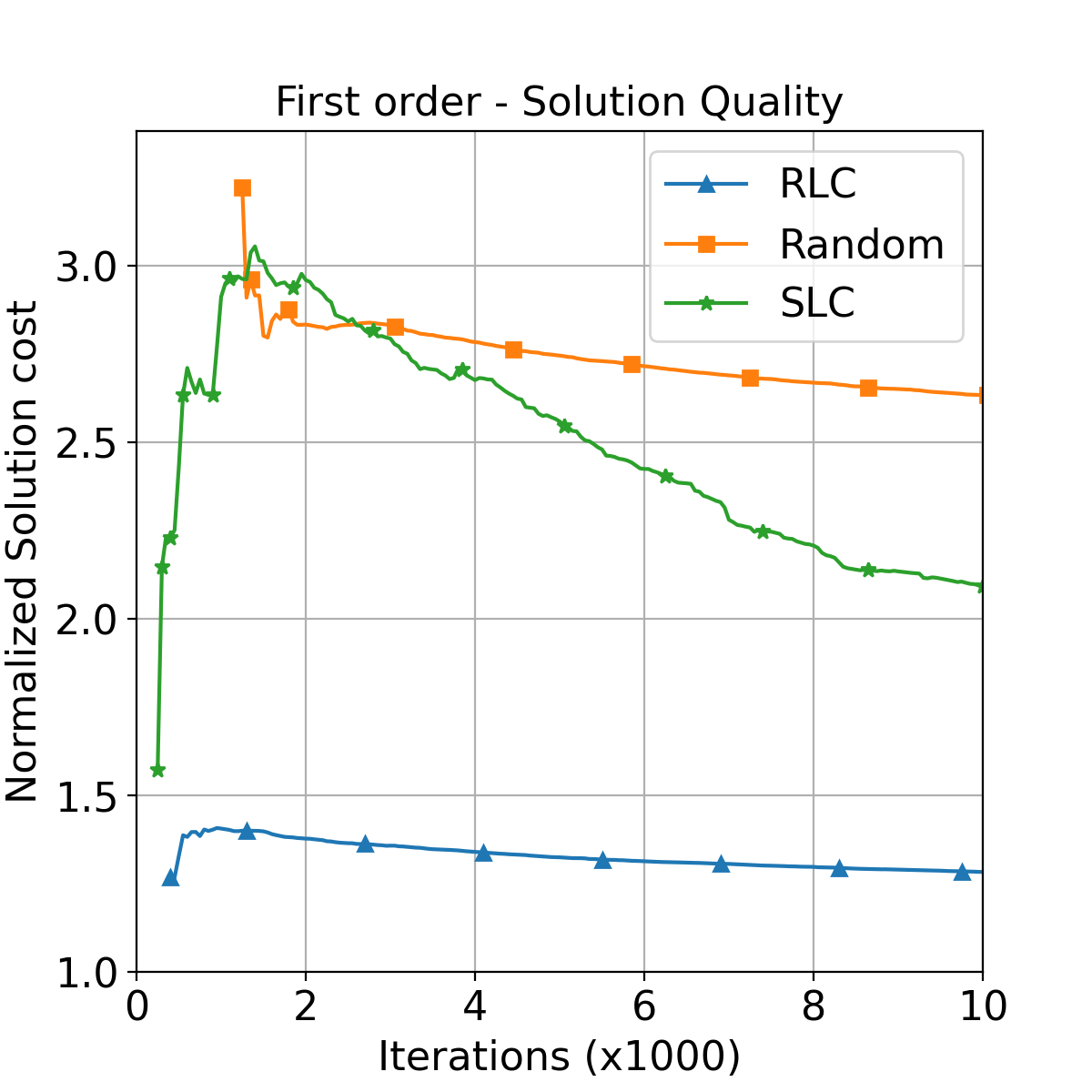}
         \vspace{-.2in}
         \caption{Results for first-order differential drive system in the cost map environments.}
         \label{fig:planner_analysis_costmaps}
\end{subfigure}\\
\begin{subfigure}[b]{\textwidth}
         \includegraphics[width=.245\textwidth]{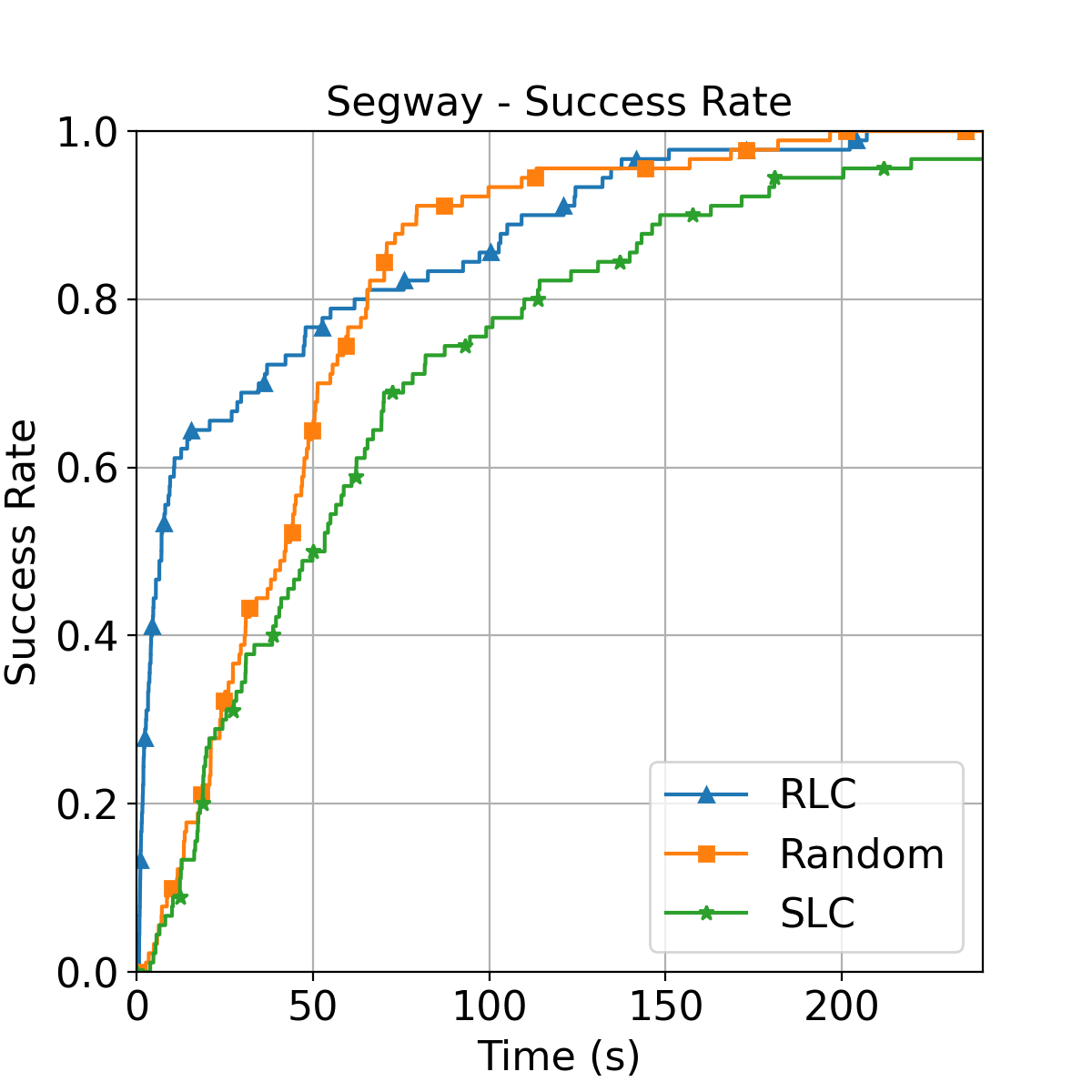}
         \includegraphics[width=.245\textwidth]{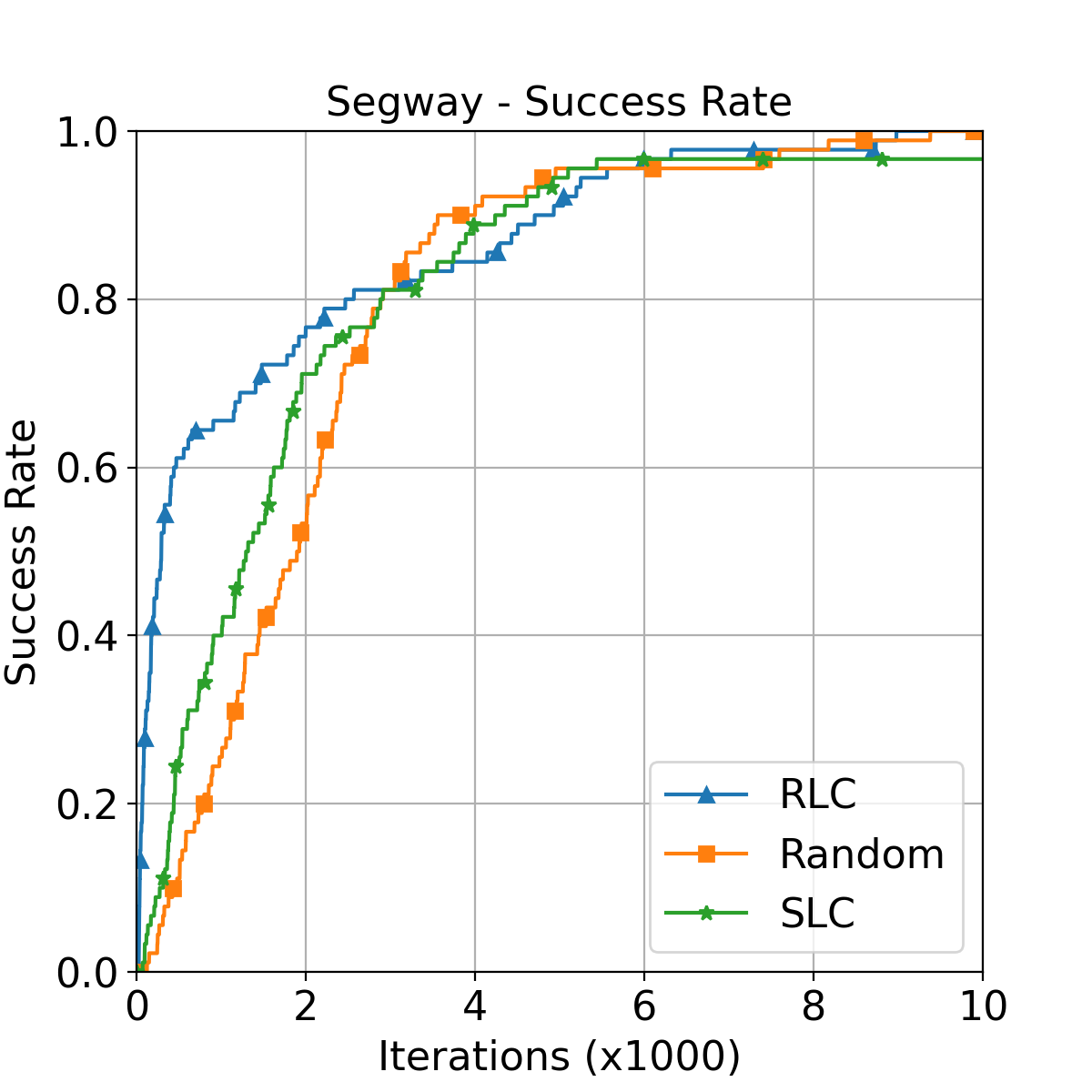}
         \includegraphics[width=.245\textwidth]{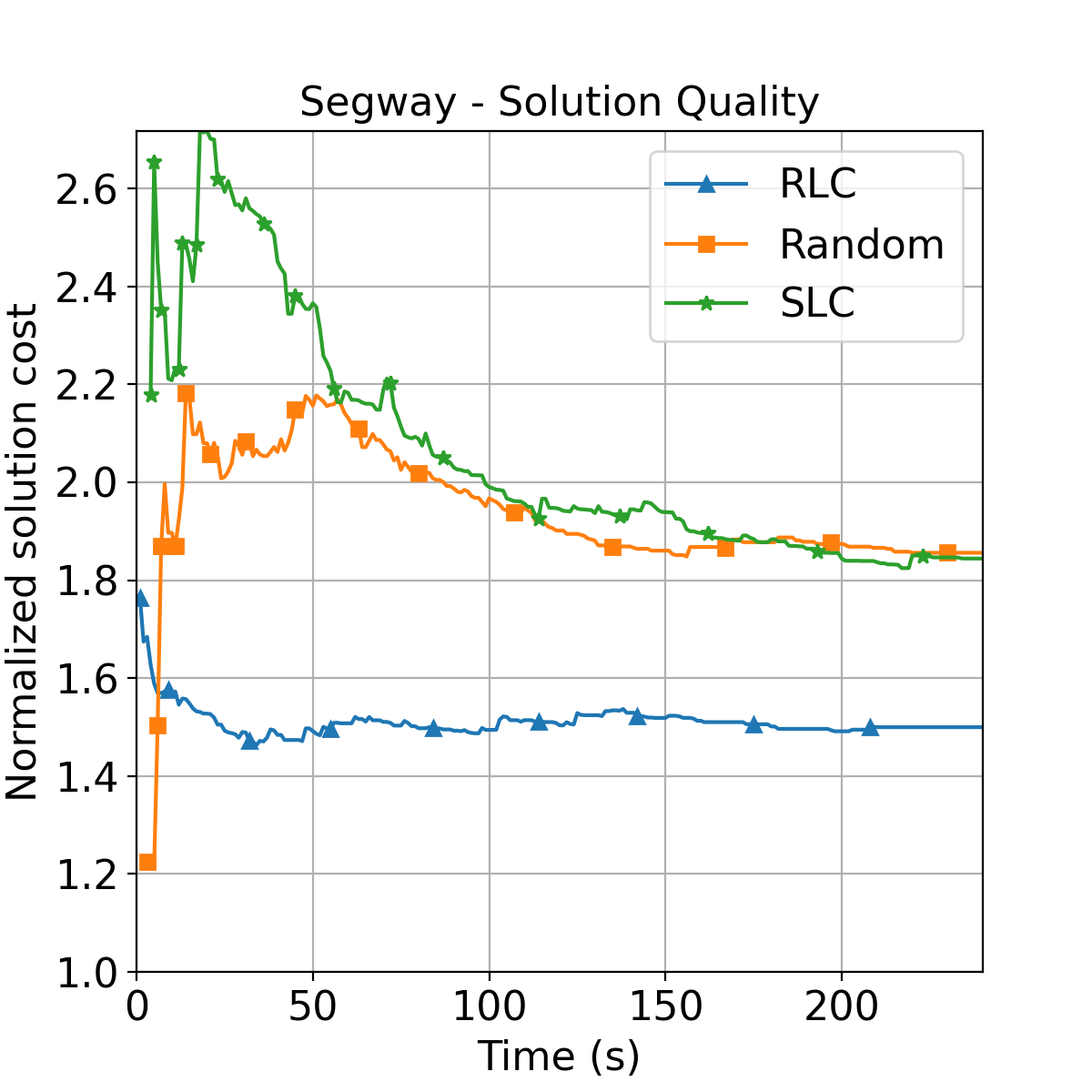}
         \includegraphics[width=.245\textwidth]{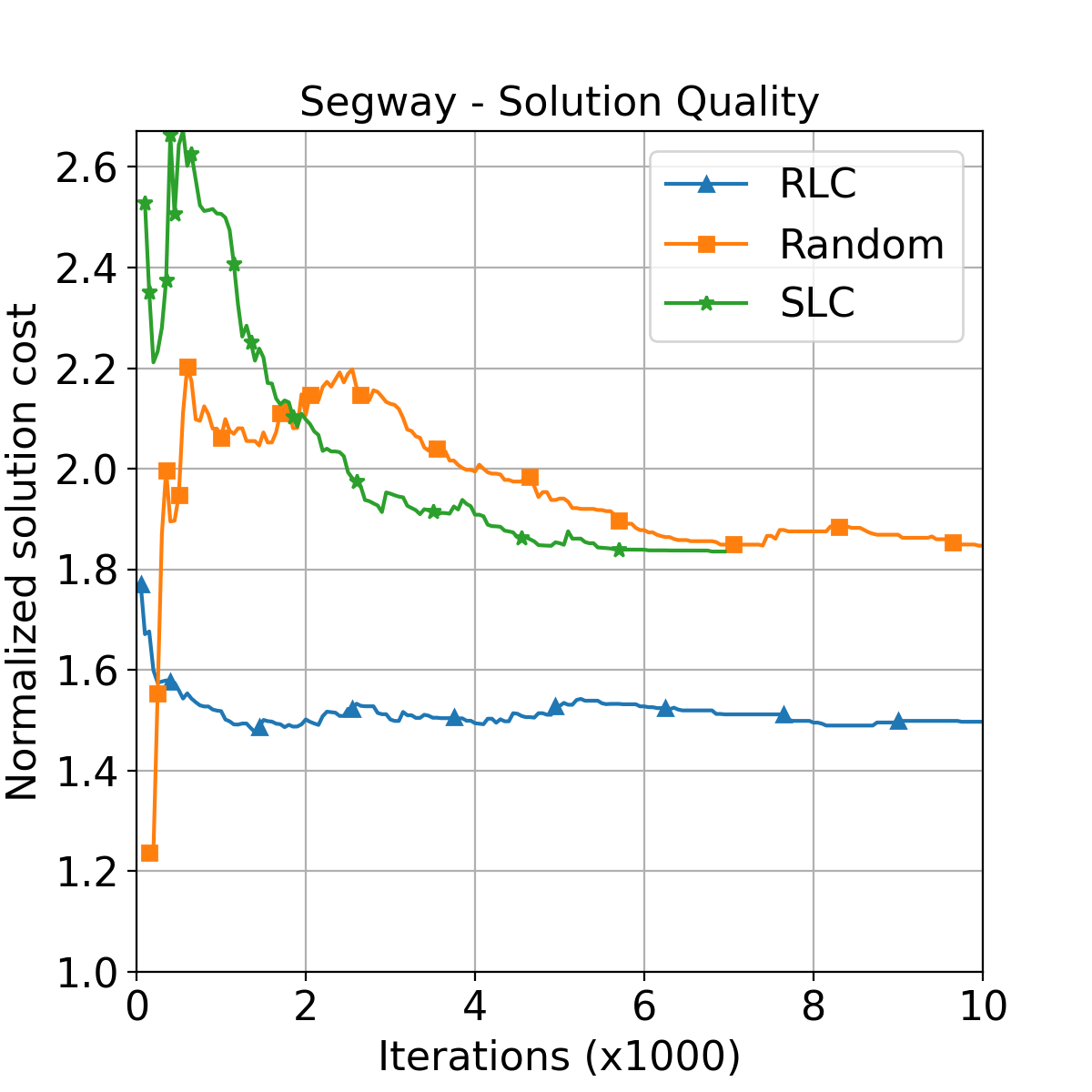}
         \vspace{-.2in}
         \caption{Results for Bullet-simulated Segway system.}
         \label{fig:planner_analysis_segway}
\end{subfigure}\\
\caption{For left and middle-left graphs, higher success rate at earlier time/iteration is better. For right and middle-right graphs, lower path cost is better. Each problem instance is run 30 times to account for different random seeds.}
\vspace{-5mm}
\end{figure*}

\subsection{Planner integration evaluation}

The integrated planning and learning solution is evaluated with three different expansion functions using a Blossom expansion of five controls each. (a) \textbf{\texttt{Random}} uses a blossom expansion of random controls in $\mathbb{U}$. (b) \textbf{\texttt{SLC}} uses the supervised learning process with a local goal provided by $\phi$ as input for the first control; for subsequent controls, a random local goal in $\mathcal{X}$ is provided as input to the supervised learned controller. (c) \textbf{\texttt{RLC}} uses SAC with local goal provided by $\phi$ as input for the first control; for subsequent controls, a random local goal in $\mathbb{X}$ is provided as input to SAC. For the Segway system, since each forward propagation is expensive, a Blossom number of 1 is used.

Across the experiments, the following metrics are measured for every planner: (1) Average normalized cost of solutions found over time/number of planner iterations, and (2) Ratio of experiments for which solution was found over time/number of planner iterations. To account for the difficulty of different planning problems, path costs are normalized by dividing by the best path cost found for a problem across any planner. For the non-cost map problems, the path cost is defined to be the duration of the solution plan. For the cost map problems, the path cost is computed according to Equation~\ref{eq:cmap_cost} with $K = 4.68$. 

For the first and second-order systems, the planner performance is evaluated on three maps from the city benchmarks \cite{sturtevant2012benchmarks}. Planner performance with the first-order system is also evaluated on three cost maps from \cite{jaillet2008transition}. For each map, ten different problem instances are generated by sampling collision-free start and goal states that have a minimum separation of $M$ pixels. For the Segway robot, three environments are defined and planner performance is evaluated across all. Example solutions found by \texttt{RLC} are given in Fig~\ref{fig:introduction}. For the city environments, the search trees are visualized.

In the city environments, \texttt{RLC} finds the lowest cost solutions very quickly for both first (Fig~\ref{fig:planner_analysis_fo}) and second-order (Fig~\ref{fig:planner_analysis_so}) systems. For the first-order system, \texttt{RLC}  finds solutions in fewer iterations than \texttt{Random} or \texttt{SLC}. For the second-order system, \texttt{SLC} finds solutions in fewer iterations than \texttt{RLC}, but \texttt{RLC} finds better quality solutions. In the cost map environments (Fig~\ref{fig:planner_analysis_costmaps}), \texttt{RLC} finds the lowest cost solutions quickly. For the same time budget, \texttt{RLC} and \texttt{Random} find solutions to more problems compared to \texttt{SLC}, while \texttt{RLC} does so in fewer iterations than \texttt{Random}. For the Segway (Fig~\ref{fig:planner_analysis_segway}), \texttt{RLC} finds lower cost solutions than both \texttt{Random} and \texttt{SLC}, and also finds more solutions in fewer iterations / lesser time. Across experiments, both \texttt{RLC} and \texttt{SLC} achieve a high success rate in fewer iterations than \texttt{Random}. The solutions obtained by \texttt{RLC}, however,  are of significantly lower cost than \texttt{SLC} and \texttt{Random}.


\begin{figure*}[ht!]
\centering
\begin{subfigure}[b]{.32\textwidth}
        \includegraphics[width=.49\textwidth]{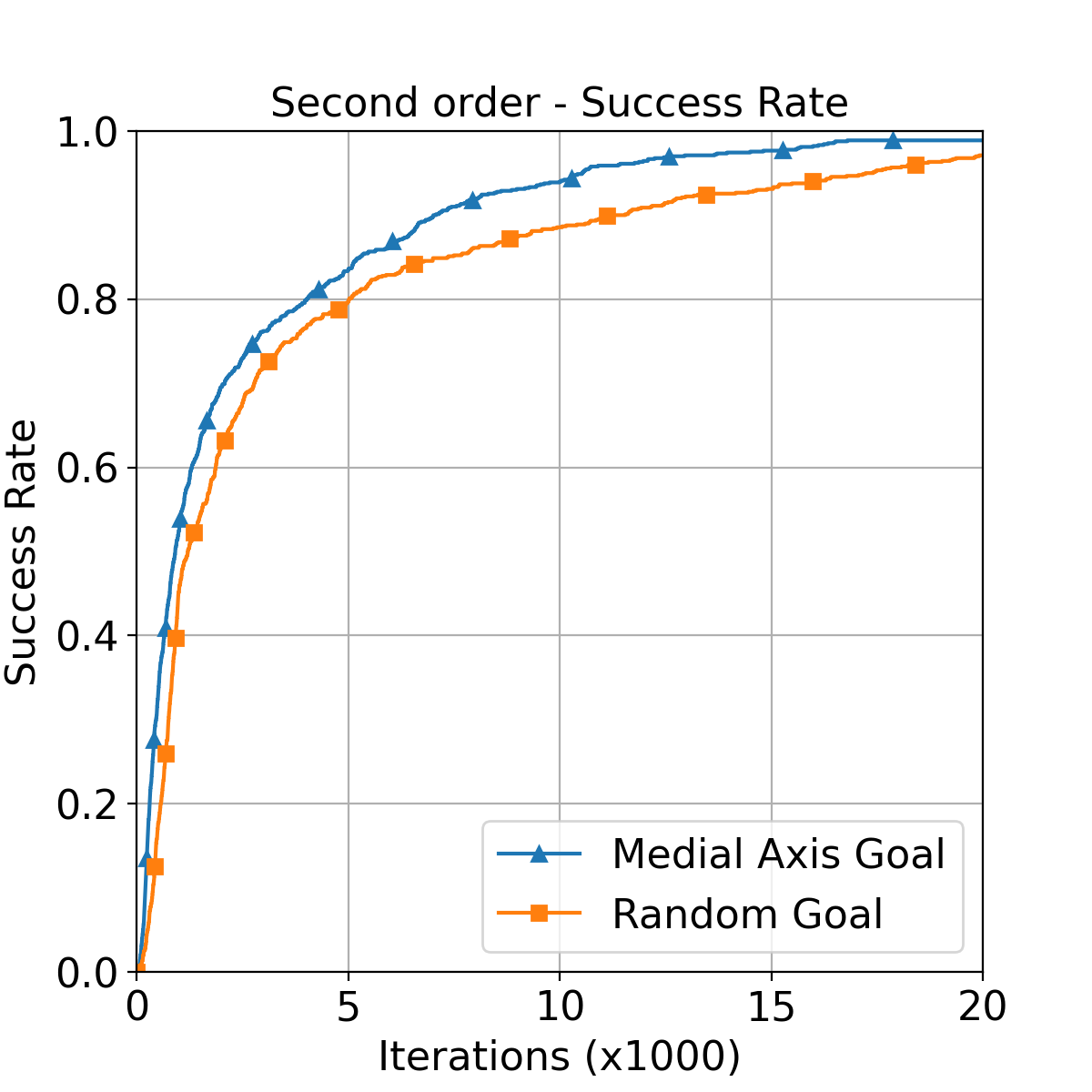}
         \includegraphics[width=.49\textwidth]{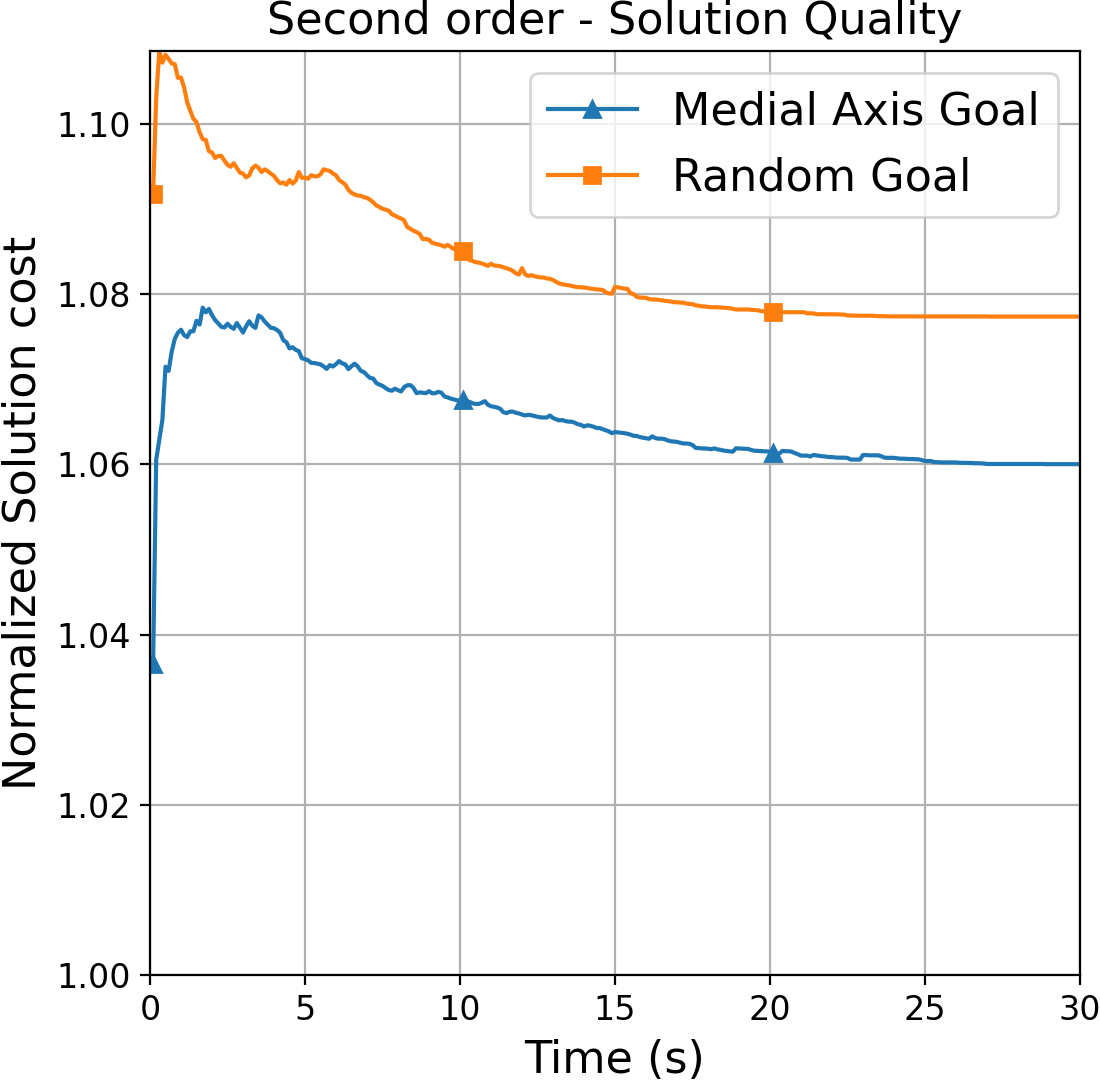}
         \caption{}
         \label{fig:ablation1-results}
\end{subfigure}
\rulesep
\begin{subfigure}[b]{.32\textwidth}
         \includegraphics[width=.49\textwidth]{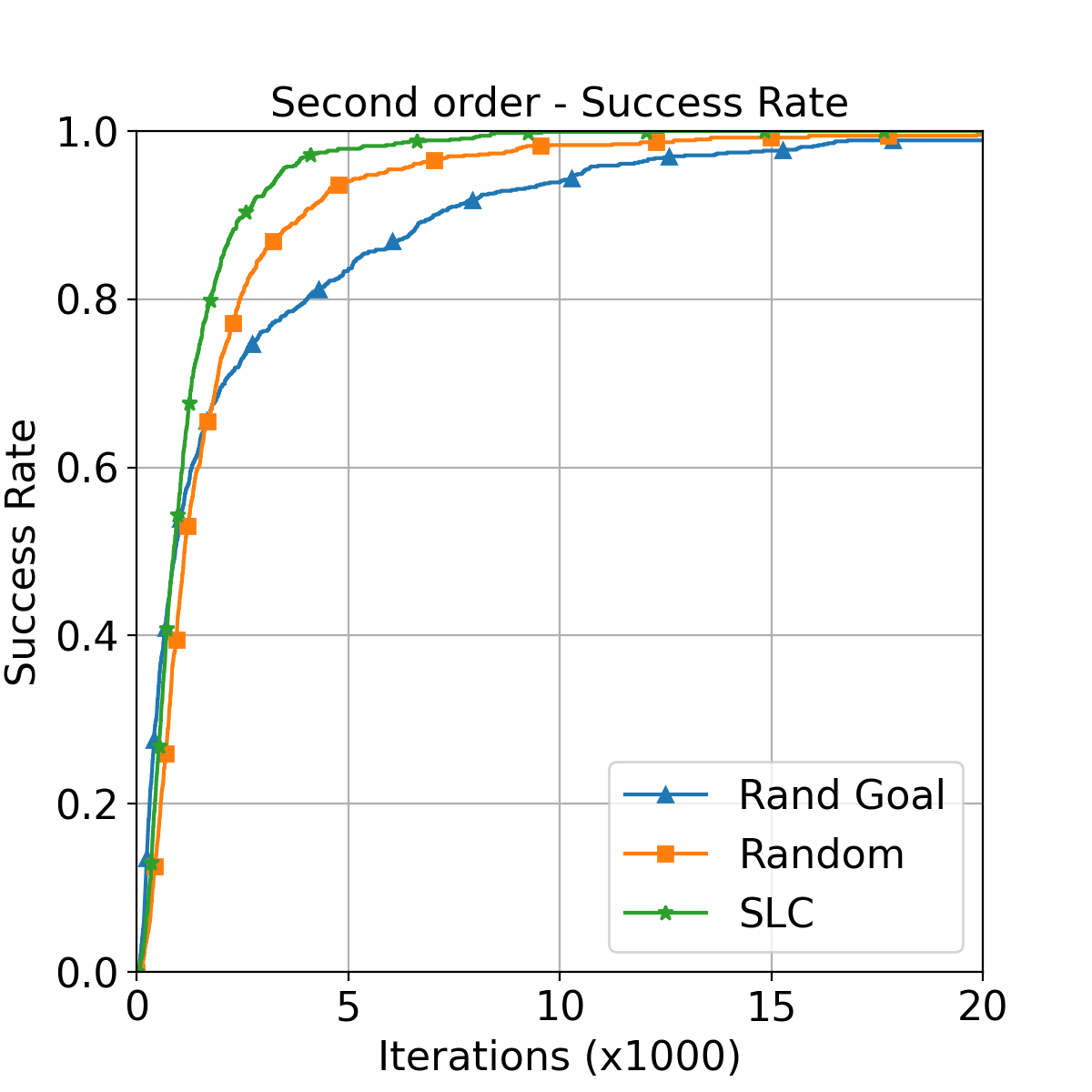}
         \includegraphics[width=.49\textwidth]{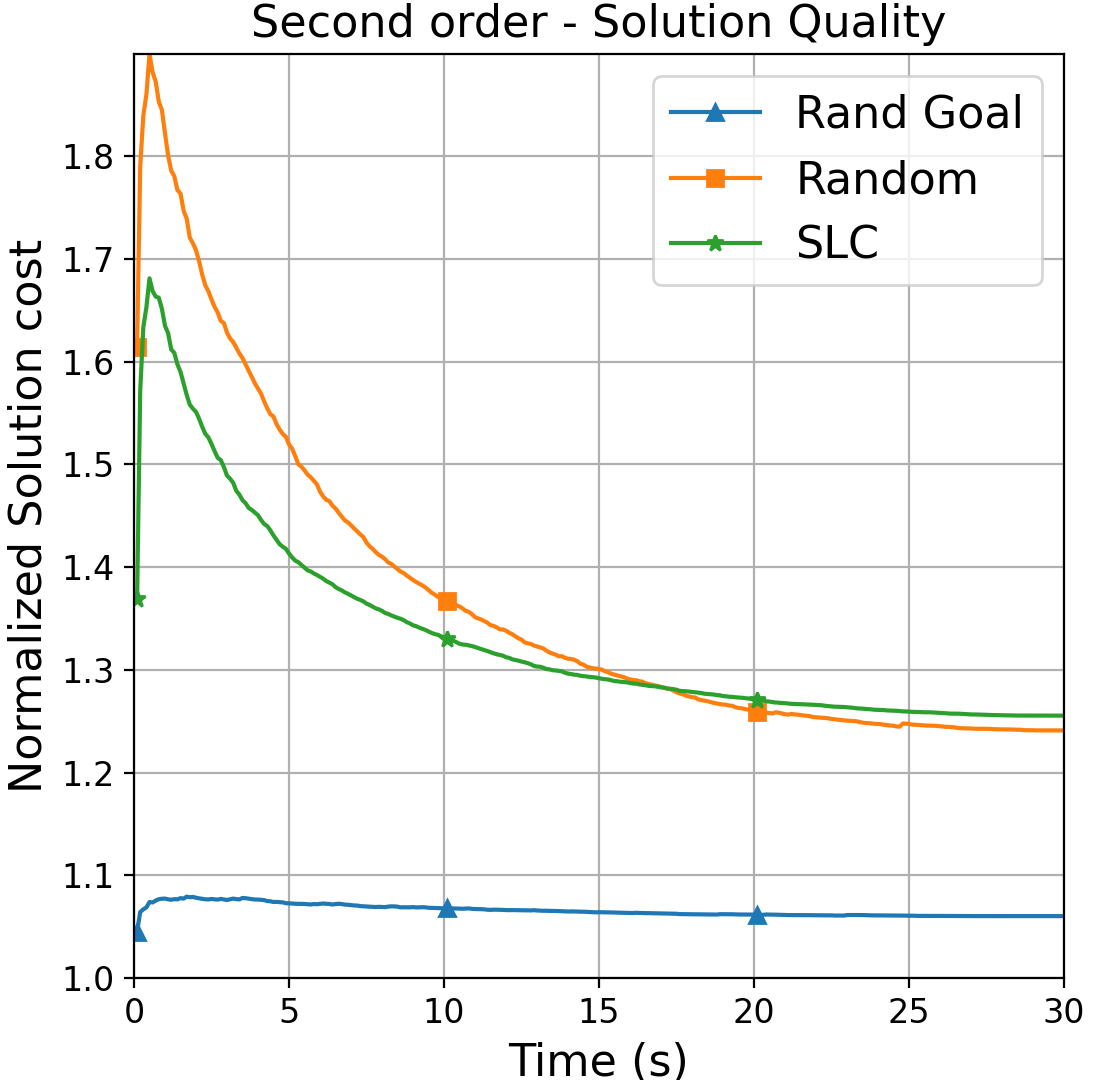}
         \caption{}
         \label{fig:ablation2-results}
\end{subfigure}
\rulesep
\begin{subfigure}[b]{.32\textwidth}
        \includegraphics[width=.49\textwidth]{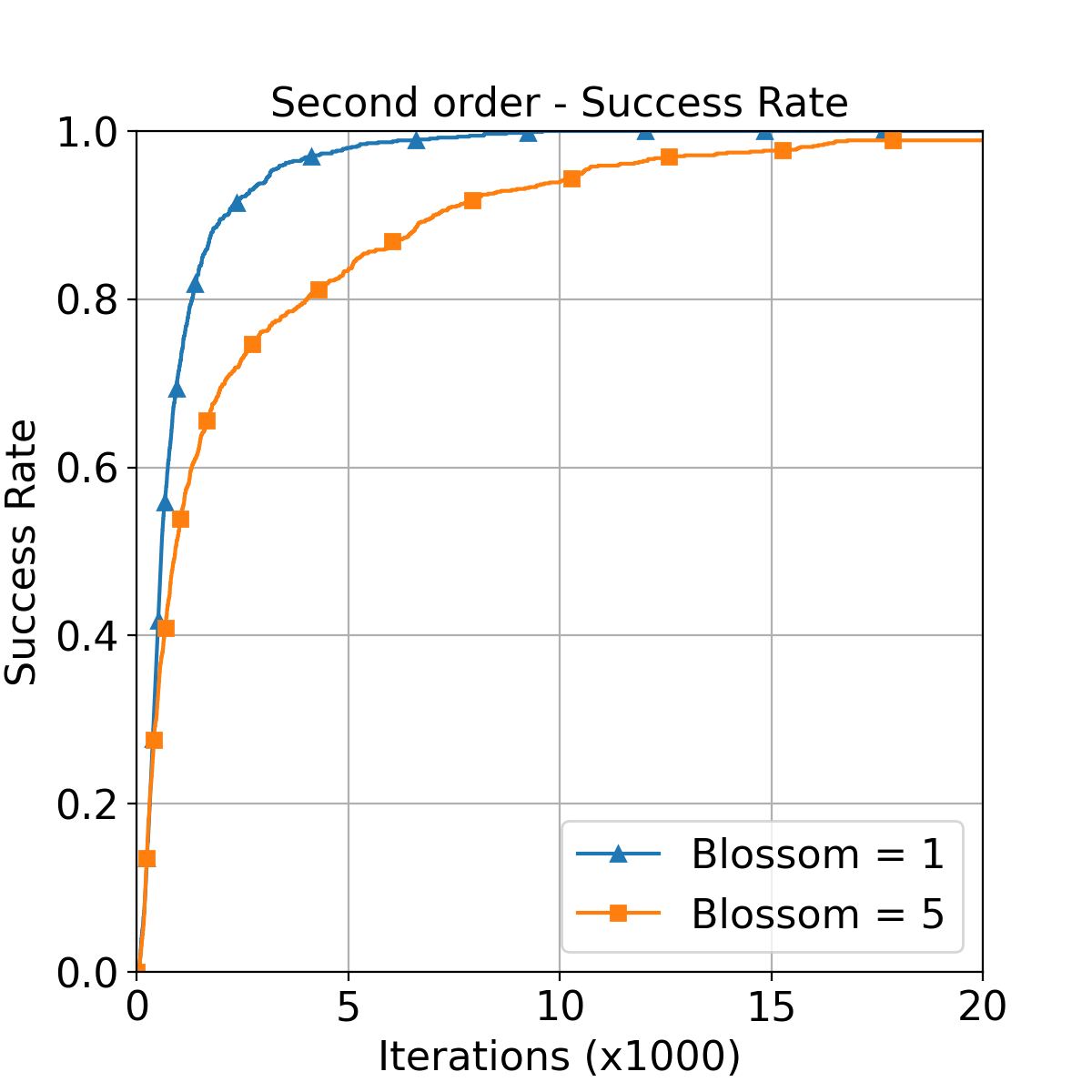}
         \includegraphics[width=.49\textwidth]{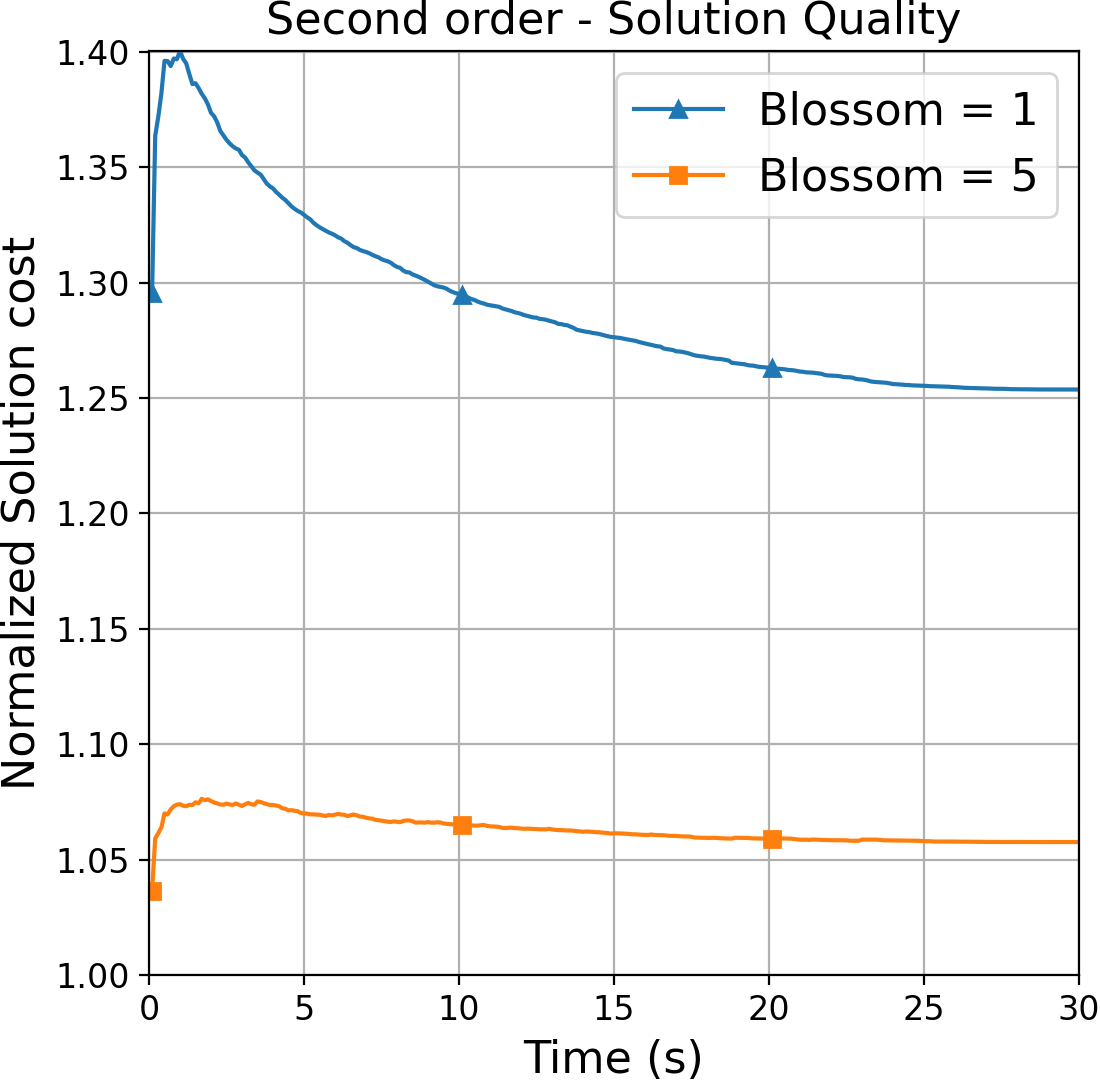}
         \caption{}
         \label{fig:ablation3-results}
\end{subfigure}
\caption{Results of the ablation experiments for the second-order system.}
\vspace{-5mm}
\end{figure*}

\subsection{Ablation Studies}

These experiments evaluate various choices for the planning process in the case of the second-order vehicle: (a) should the first propagation be informed given medial axis information, (b) should consecutive propagations use random controls (\texttt{Random}), or use a random local goal and supervised learning (\texttt{SLC}) or use a random local goal and SAC (\texttt{Rand Goal}), and (c) what is a good blossom number. The planner using \texttt{Medial Axis Goal} finds lower cost solutions in fewer iterations than the one using \texttt{Random Goal} (Fig.~\ref{fig:ablation1-results}). The planner using \texttt{SLC} finds solutions in fewer iterations than  using \texttt{Rand Goal} but \texttt{RandGoal} finds higher quality solutions overall (Fig.~\ref{fig:ablation2-results}). \texttt{RandGoal} also remains competitive in terms of the number of solutions found compared to \texttt{Random}. A smaller Blossom number finds solutions faster (Fig~\ref{fig:ablation3-results}). The quality of solutions found, however, improves considerably when a higher Blossom number is used. The best-performing choices across these experiments are used for \texttt{RLC} evaluated in Section~\ref{sec:results}-B.

%% file: sections/06_conclusion.tex
\section{Conclusion}
\label{sec:conclusion}

This paper proposes a learning-based node expansion method that improves the path quality and computational efficiency of sampling-based kinodynamic planners for vehicular systems. It consists of a learned controller that can be trained with small data requirements for different robotic systems, and a selection process of local goal states that is computed once for every planning problem. 

The inference time of the learned controller can further be improved by storing it upon training to an integrated circuit, such as an FPGA. Furthermore, the local goal state selection can also be learned in conjunction with the controller. Similarly, the proposed method can also be integrated with other learning primitives for planning, such as a learned selection procedure. Next steps include simulations with more complex systems and experiments on real vehicular systems.